\documentclass[lettersize,journal]{IEEEtran}
\usepackage{amsmath,amsfonts}
\usepackage{array}
\usepackage[caption=false,font=footnotesize,labelfont=sf,textfont=sf]{subfig}
\usepackage{textcomp}
\usepackage{stfloats}
\usepackage{url}
\usepackage{verbatim}
\usepackage{graphicx}
\usepackage{bm} %符号加粗
\usepackage{xcolor}    % 用于设置颜色
\usepackage{cite}  % 这个符号能把参考文献连起来
\usepackage{multirow}
\usepackage{algorithm}
\usepackage{algpseudocode} 

\usepackage{amsmath}  
\usepackage{hyperref}
\usepackage{booktabs}

\usepackage{makecell}

\def\BibTeX{{\rm B\kern-.05em{\sc i\kern-.025em b}\kern-.08em
    T\kern-.1667em\lower.7ex\hbox{E}\kern-.125emX}}
\usepackage{balance}

\begin{document}
\title{\fontsize{22}{28}\selectfont Transferable Deployment of Semantic Edge Inference Systems via Unsupervised Domain Adaption}
\author{Weiqiang Jiao, Suzhi Bi, Xian Li, Cheng Guo, Hao Chen, and Zhi Quan
\thanks{Weiqiang Jiao, Suzhi Bi, Xian Li, and Zhi Quan are with the State Key Laboratory of Radio Frequency Heterogeneous Integration and College of Electronics and Information Engineering, Shenzhen University, Shenzhen, Guangdong 518060, China (e-mail: jiaoweiqiang2023@email.szu.edu.cn; bsz@szu.edu.cn; xianli@szu.edu.cn; zquan@szu.edu.cn).}
\thanks{Cheng Guo and Hao Chen are with the Peng Cheng Laboratory, Shenzhen 518066, China (e-mail: guoch03@pcl.ac.cn, chenh03@pcl.ac.cn).}
}

\maketitle

\begin{abstract}
This paper investigates deploying semantic edge inference systems for performing a common image clarification task. In particular, each system consists of multiple Internet of Things (IoT) devices that first locally encode the sensing data into semantic features and then transmit them to an edge server for subsequent data fusion and task inference. The inference accuracy is determined by efficient training of the feature encoder/decoder using labeled data samples. Due to the difference in sensing data and communication channel distributions, deploying the system in a new environment may induce high costs in annotating data labels and re-training the encoder/decoder models. To achieve cost-effective transferable system deployment, we propose an efficient Domain Adaptation method for Semantic Edge INference systems (DASEIN) that can maintain high inference accuracy in a new environment without the need for labeled samples. Specifically, DASEIN exploits the task-relevant data correlation between different deployment scenarios by leveraging the techniques of unsupervised domain adaptation and knowledge distillation. It devises an efficient two-step adaptation procedure that sequentially aligns the data distributions and adapts to the channel variations. Numerical results show that, under a substantial change in sensing data distributions, the proposed DASEIN outperforms the best-performing benchmark method by 7.09$\%$ and 21.33$\%$ in inference accuracy when the new environment has similar or 25 dB lower channel signal to noise power ratios (SNRs), respectively. This verifies the effectiveness of the proposed method in adapting both data and channel distributions in practical transfer deployment applications.
\end{abstract}


\section{Introduction}\label{I}
\IEEEPARstart{T}hanks to the advancement of artificial intelligence (AI), it becomes prevalent in recent years to deploy smart Internet of Things (IoT) systems using deep neural networks (DNNs) to perform complex inference tasks, e.g., computer vision based object recognition \cite{a1, a2, a3}. In particular, wireless IoT devices, such as video surveillance cameras, are systematically deployed at target locations to collect real-time sensing data and collaboratively accomplish specific inference tasks. The performance of on-device AI inference, however, is significantly constrained by the limited battery energy and computing power of IoT devices. To prolong the battery lifetime and improve the inference performance, a promising approach is to offload the collected sensing data to nearby mobile edge computing (MEC) servers of much stronger computing power \cite{a6, a7, a7_1, a7_2, a7_3}. In an industrial IoT application, for example, MEC servers can flexibly assign real-time computation offloading tasks to available computing entities for reduced execution delay \cite{a13_1, a13_2, a13_3}. Nevertheless, simultaneous wireless transmissions of a substantial amount of raw data (such as images or videos) can lead to high communication delay. The recent development of task-oriented semantic communication technology offers an effective means to address this challenge \cite{a8, a9, a9_1, a9_2, a9_3}. Specifically, task-oriented semantic communication extracts and transmits only the substantive semantic information relevant to the inference task contained in the raw sensing data. As such, it allows for a significant reduction of communication data size and thus more efficient data delivery under bandwidth constraints. Several studies have developed semantic communication techniques based on DNNs to send information like images \cite{a10}, videos \cite{a11}, speech \cite{a12}, and text \cite{a13}. Leveraging the techniques of semantic communication, the IoT devices first locally encode the sensing data into semantic features and then transmit it to edge servers for subsequent data fusion and task inference.

\begin{figure}[t]
	\centering
	\subfloat[Scenario1]{\includegraphics[width=0.49\hsize,height=0.38\hsize]{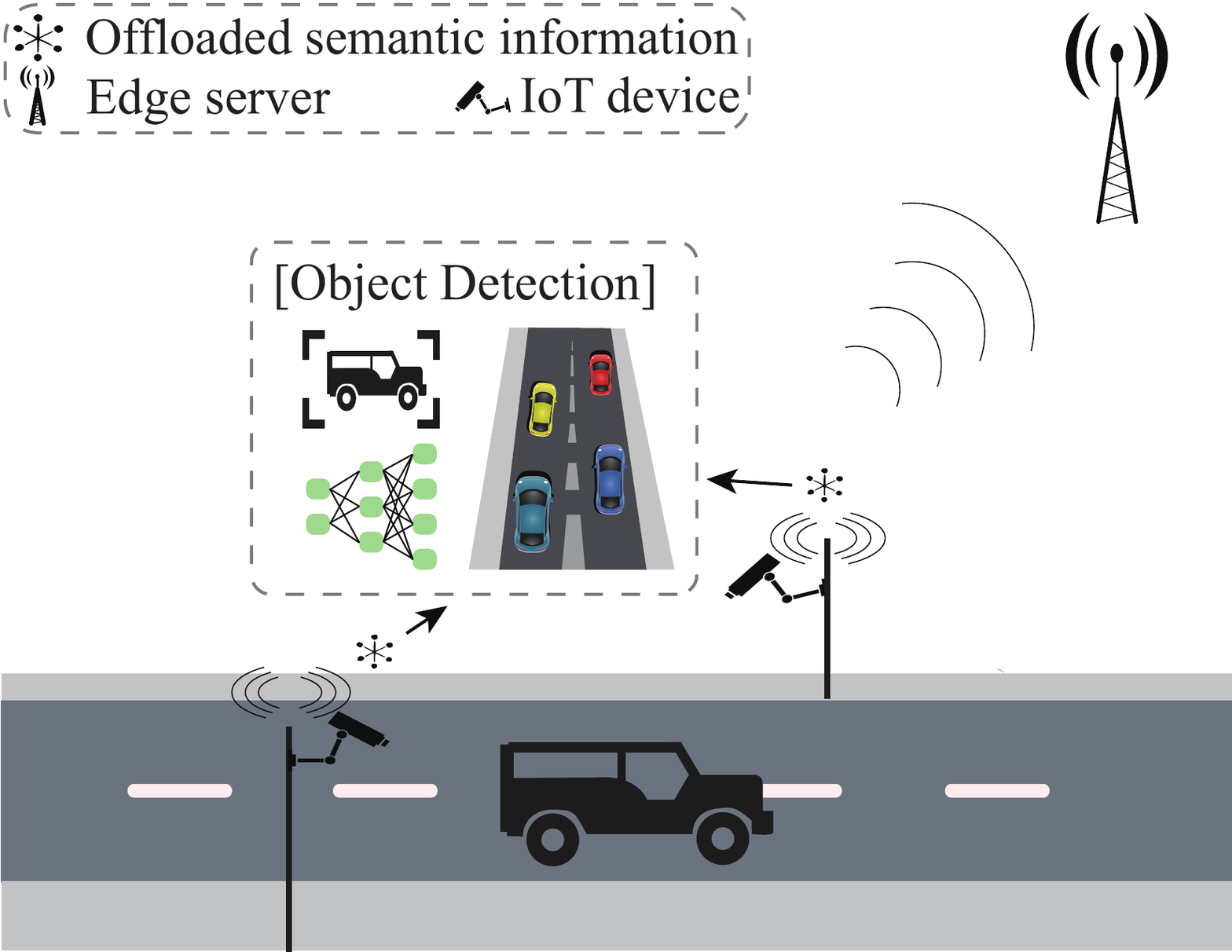}\label{scenario1}}
	\hfill
	\subfloat[Scenario2]{\includegraphics[width=0.49\hsize,height=0.32\hsize]{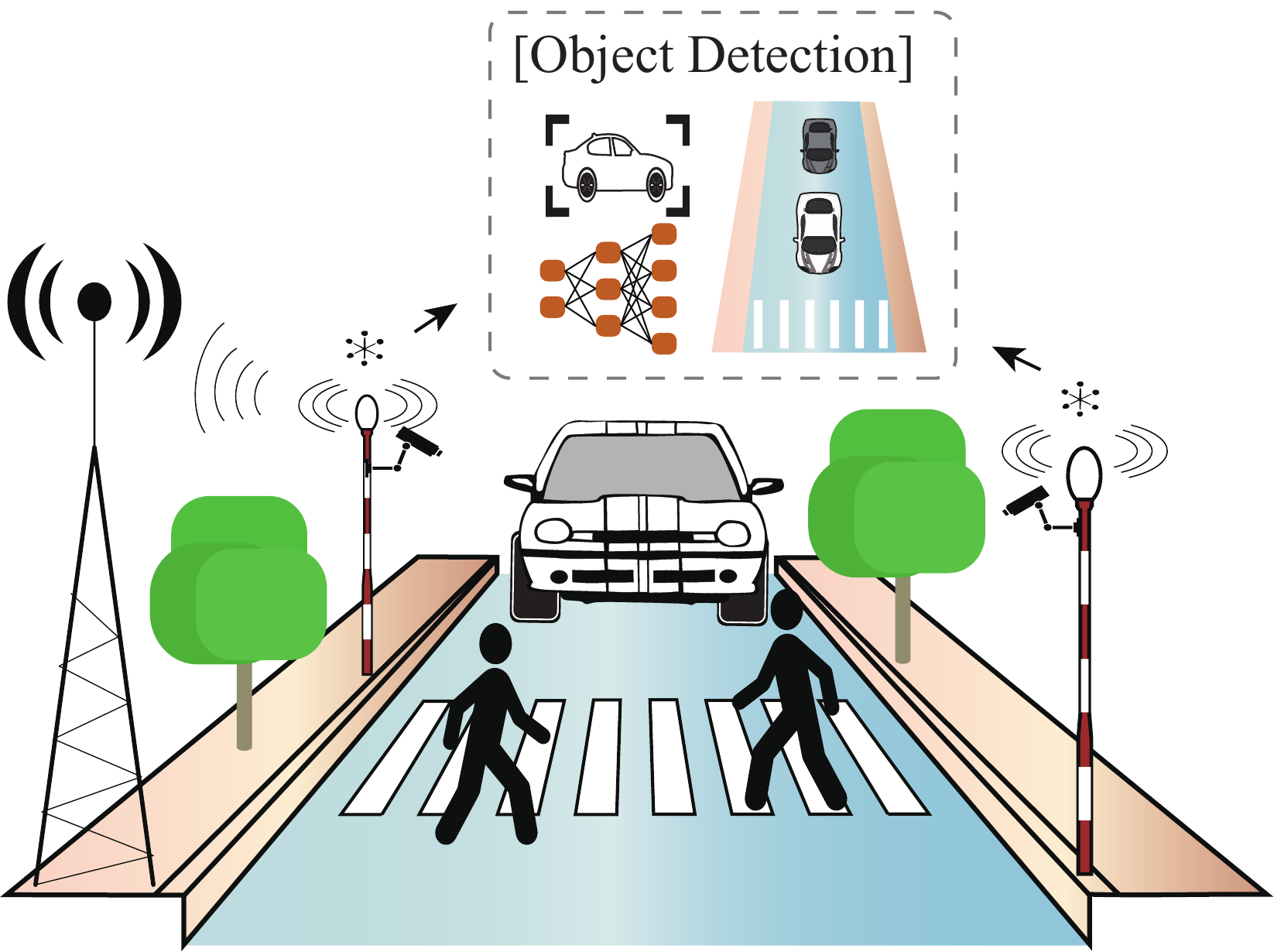}\label{scenario2}}
	\caption{Two distinct deployment scenarios of semantic edge inference system for object detection.}
	\label{fig: room}
\end{figure}

Despite its potential performance advantage, deep learning-based semantic communication is known susceptible to dynamic wireless environments. That is, the encoder/decoder trained using the data samples collected under a particular channel distribution (e.g., at a given signal to noise power ratio (SNR)), needs to be re-trained when the channel distribution changes significantly (e.g., due to blockage) to maintain satisfactory inference accuracy. To improve the adaptation of DNN-based models to varying channel conditions, Beck \emph{et al}. \cite{a14} proposed to train separate DNN models under different SNRs offline and switch to the one that most matches the real-time SNR. This SNR-based model switching method, however, requires collecting a huge number of training data samples to adapt large SNR variations. Another method is to embed SNR as soft information input to the DNN encoder or decoder models. For example, Ding \emph{et al}. \cite{a15} introduced a joint source and channel coding (JSCC) scheme with SNR adaptation ability, wherein the decoder estimates the SNR and inputs it as a feature to aid in the decoding process. Xu \emph{et al}. \cite{a16} considered SNR as input information design its DNN structure, which allocates a higher code length for more reliable data transmission under low SNR, and vice versa. Shao \emph{et al}. \cite{a17} proposed a variable-length feature encoding scheme based on dynamic neural networks, which can adaptively adjust the activation dimensions of the coded features according to different channel conditions. \par

Besides the variation of channel conditions, the deployment of semantic edge inference system in different locations and scenarios also induces variation of the sensing data distribution. Consider an example traffic monitoring application in Fig. \ref{fig: room}, two sets of IoT cameras are deployed at different locations, such that the distributions of the collected video sensing data differ by the background environments. To address the joint variation of data and channel distributions, a naive approach is to train two sets of DNN encoders/decoders by collecting sufficient labeled data samples in the different deployment scenarios. However, it is time-consuming and labor-intensive when deploying a large number of edge inference systems. Essentially, it fails to capitalize on the task-relevant semantic similarity in different deployment scenarios. Recent advance in transfer learning provides an effective means to reuse DNN models to accomplish related tasks, which significantly reduces the online training time and improves the sample efficiency \cite{a18, a19}. In general, it involves an offline training stage in a familiar ``source domain'' and an online training stage in an unfamiliar ``target domain''. Depending on the availability of labeled data samples in the target domain, few-shot learning \cite{a20, a21} and domain adaptation \cite{a22, a23, a24, a25} are two common transfer learning methodologies. \par

Some recent studies have applied transfer learning to semantic communication. For instance, Xie \emph{et al}. \cite{a26} proposed a deep learning-based semantic communication system (DeepSc) for text transmission. To adapt to different communication environments, they froze the pre-trained semantic encoder/decoder layers in stable channel environments, while retraining the channel encoder/decoder when channel environment fluctuate. Feng \emph{et al}. \cite{a27} applied the few-shot learning Model-Agnostic Meta Learning (MAML) algorithm to address the problem of varying data distributions in end-to-end semantic communication systems used for image transmission. However, this requires fine-tuning the encoder network using a small number of labeled data samples from the target domain. When the labeled target domain data samples are unavailable or costly to obtain, Sun \emph{et al}. \cite{a281} proposed an SKB-based lightweight multi-level feature extractor, which includes an intermediate feature extractor, visual autoencoder, and semantic autoencoder. The multi-level encoding approach facilitates extracting domain-invariant features while supporting zero-shot learning. Zhang \emph{et al}. \cite{a28} proposed a domain adaptation approach for image transmission with data distribution variations. They designed a GAN (general adversarial networks)-based method to convert the observed data samples in the target domain into a form that the existing DNN models can efficiently process without further training. However, the training of GAN-based DNNs is computationally expensive and requires a large number of target domain data samples. In the practical deployment of semantic edge inference system, the joint variation of channel and data distributions leads to a higher level of complexity in online data collection and training. How to swiftly adapt to simultaneous variations of both factors remains a challenging problem. \par

In this paper, we propose an efficient Domain Adaptation method for Semantic Edge INference systems (DASEIN) to achieve efficient and transferable system deployment. Specifically, starting with a semantic edge inference system (i.e., the source domain) with sufficient labeled data samples, we intend to maintain high inference accuracy when extending the deployment in an unfamiliar scenario (target domain) without any labeled sample. The detailed contributions of the paper are as follows:

\begin{itemize}
	\item[1)]
	\textbf{Unsupervised Adaptation to Joint Data-Channel Variation}: We propose a transferable deployment method of the semantic edge inference system DASEIN, which adaptively accommodates changes in joint data-channel variation. DASEIN exploits correlations between source and target domains without requiring annotations in the target domain, enabling swift and cost-effective deployment in dynamic scenarios.
	\item[2)]
	\textbf{Efficient Two-step Adaptation Procedure}: DASEIN consists of two sequential steps. First, it applies an unsupervised domain adaptation (UDA) method with a warm-up technique to train a DNN model, such that it can adapt to the data distribution shifts in the target domain under constant SNR. By treating the obtained DNN model as the teacher model, it then applies knowledge distillation (KD) with an unreliable sample filtering mechanism to train a student model adaptable to the actual SNR in the target domain.
	\item[3)]
	\textbf{Design Insight in Transfer Deployment}: Numerical results show that, under a substantial change in sensing data distributions, the proposed DASEIN outperforms the best-performing benchmark method by 7.09$\%$ in inference accuracy when the target and the source domains have similar channel SNRs. Meanwhile, the performance advantage increases to 21.33$\%$ when the channels of the target domain are 25 dB worse. Nonetheless, to achieve excellent inference accuracy under very low SNR (e.g., -20 dB), it is necessary to increase the transmit feature dimension in addition to using the model adaptation method proposed.
	\item[4)]
	\textbf{Digital Implementation Method}: We extend the application of the proposed DASEIN to digital communication systems, where the features are transmitted in digital signals. In particular, we design a continuous surrogate function to address the non-differentiable quantization problem in model training. Numerical results show that the digital scheme in general achieves higher inference accuracy than its analog counterpart, thanks to the noise-resilient capability of digital modulations.
\end{itemize}\par

The rest of the paper is organized as follows. We model the semantic edge inference system in Section \ref{II} and introduce the proposed DASEIN method in Section \ref{III}. We describe the extension to digital communications in Section \ref{III2}. In Section \ref{IV}, we evaluate the performance of DASEIN and conclude the paper in Section \ref{V}.

\section{System Model}\label{II}
\begin{table}[t]
	\centering
	\caption{NOMENCLATURE}
	\small 
	\renewcommand{\arraystretch}{1.4} % 设置行高为1.5倍
	\begin{tabular}{p{1.0cm}p{7.0cm}}
		\toprule
		Symbol      & Description. \\
		\hline
		$ Ch, W, H $               & The number of channels, height, and width of sample. \\
		$\mathcal{S},\mathcal{T}$  & Source domain and target domain. \\
		$ D^* $                    & The dataset of $ * $ domains. \\
		$ x, y $                   & The data collected by devices and its ground-truth. \\
		$ \hat{\bm{p}}, \hat{\bm{p}}_{max} $   & The output of decoder $ D $ and maximum entry of $ \hat{\bm{p}} $. \\
		$ \hat{y}, \hat{y}_{oh} $  & The inference label and one-hot vector of input sample. \\
		$ \theta $                 & The parameters of the DNN model. \\
		$ k $                      & The index of devices, in total $ K $ devices. \\
		$ f^k $                    & Semantic feature generated by $ E_{SRE}^k $. \\
		$ z^k, \hat{z}^k $         & Compressed semantic feature and noisy semantic feature received at device $ k $. \\
		$ \hat{z} $                & The concatenated noisy semantic feature. \\
		$ tc, st $                 & Teacher and student models. \\
		$ \omega_i^c $             & The weights of samples $ x_i $  belonging to the class c. \\
		$ z_q^k, \hat{z}_q^k $     & The digital signal and noisy digital signal at device $ k $. \\
		$ z_m^k, \hat{z}_m^k $     & The modulated symbol and noisy modulated symbol at device $ k $. \\\hline
	\end{tabular}
	\label{tab:NOMENCLATURE}
\end{table}

\begin{figure}[t]%调节图片位置，h：浮动；t：顶部；b:底部；p：当前位置
	\centering
	\includegraphics[width=9cm]{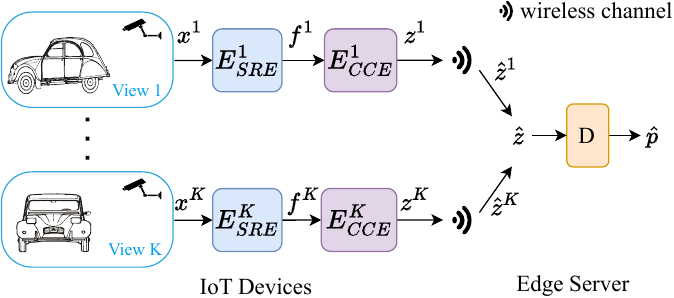}% 中括号中的为调节图片大小
	\caption{The schematics of the considered edge inference system.}
	\label{2diagram}%文中引用该图片代号
\end{figure}

As depicted in Fig. \ref{2diagram}, we consider an edge computing system consisting of $ K $ IoT devices (cameras) and an edge server. In particular, the IoT devices collaboratively monitor a target of interest from different viewpoints. Each IoT device locally encodes its observed image information into a semantic feature, and sends to the edge server, which then combines the $ K $ features and derives a final inference result, e.g., a ``black car''. For ease of reference, we list the symbols used in this paper in Table \ref{tab:NOMENCLATURE}. \par

\subsection{Edge Inference Model\label{II-A}}
\subsubsection{IoT Devices\label{II-A1}}
The IoT devices are designed to collaboratively extract semantic features relevant to the inference task from the captured images. The $ K $ devices are indexed by $ k\in \{1,2,\cdots,K\} $, and we denote the captured image at the device $ k $ as $ x^k \in \mathbb{R} ^{Ch \times H \times W} $, where $ Ch $, $ H $, and $ W $ denote the number of channels, height, and width of this image sample, respectively. Let $ y \in \mathcal{Y} $ represent the ground-truth corresponding to the image $ x $. The set $ \mathcal{Y} $ encompasses categories such as ``car'', ``bike'', and others, indicative of the object depicted in the image. $ E_{SRE}^k(\cdot, \theta_{SRE}) $ parameterized by $ \theta_{SRE} $ denotes the semantic representation extractor, which produces the semantic feature $ f^k \in \mathbb{R} ^{a_{in} \times 1} $ as:
\begin{equation}\label{e1}
	f^k = E_{SRE}^k(x^k, \theta_{SRE}),  k = 1,2,\cdots, K. 
\end{equation}
The semantic feature represents an abstract expression of the core meaning of image data. To further compress the data, we use a compress and channel encoder $ E_{CCE}^k(\cdot, \theta_{CCE}) $ parameterized by $ \theta_{CCE} $ to process $ f^k $ into compressed semantic feature $ z^k \in \mathbb{R} ^{a_{out} \times 1} $, i.e., 
\begin{equation}\label{e2}
	z^{k} = E_{CCE}^k(f^{k}, \theta_{CCE}),  k = 1,2,\cdots, K.
\end{equation}
Notably, $ a_{in} $ and $ a_{out} $ above respectively denote the input and output dimensions of $ E_{CCE}^k $, respectively. The IoT devices will then send compressed semantic feature $ z^{k} $ to the edge server in orthogonal wireless channels, thus concurrent transmissions will not interfere with each other. By normalizing against the channel gains, we consider additive white Gaussian noise (AWGN) wireless channels between the IoT devices and the edge server. The channel output symbol received at the receiver can be expressed as:
\begin{equation}\label{e3}
	\hat{z}^k = z^{k} + n^k, k=1,\cdots, K,
\end{equation}
where $ n^k \sim N(0, \sigma_k I) $, $\sigma_k $ denotes the normalized receiver noise power and $ I $ is an identity matrix. Notice that the value of $ \sigma_k $ is related to the channel gain between the $ k $th device and the edge server, where a larger $ \sigma_k $ corresponds to a weaker channel and vice versa. We denote $  \bm{\sigma} = \left\{\sigma_k\right\}_{k=1}^K $ to represent the wireless channel condition of the edge system. 

\subsubsection{Edge Server\label{II-A2}}
The receiver concatenates $ \hat{z}^k $ into $ \hat{z} = [\hat{z}^1, \hat{z}^2, \cdots, \hat{z}^K] $, then inputs to the decoder to obtain the predicted probability distribution $ \bm{\hat{p}} $, which can be expressed as:
\begin{equation}\label{e4}
	\bm{\hat{p}} = D(\hat{z}, \theta_{D}).
\end{equation}
Here, the decoder $ D $ is parameterized by $ \theta_{D} $, $ \bm{\hat{p}} = [\hat{p}_1, \cdots, \hat{p}_c,\cdots, \hat{p}_C] $, where $ C $ denotes the total number of categories. In the end, the edge server selects the index that corresponds to the maximum value $ \bm{\hat{p}}_{max} $ in $ \bm{\hat{p}} $ as the inference label $ \hat{y} $ of image $ x $. Accordingly, we can obtain the one-hot vector $ \hat{y}_{oh} $ by creating a zero vector of length equal to $ C $, setting the position corresponding to the label $ \hat{y} $ to 1, while the other positions are 0. We denote the entire set of DNN encoder/decoder of the system as $\theta = \left\{ \theta_{SRE}, \theta_{CCE}, \theta_D\right\}$.

\subsection{Problem Description\label{II-B}}
We explore a scenario in which the source domain contains an abundance of labeled samples. Mathematically, we define the source dataset as $ \mathcal{D}^{\mathcal{S}} = (x_i^{{\mathcal{S}}_1}, \cdots, x_i^{{\mathcal{S}}_K}, y_i^{{\mathcal{S}}})_{i=1}^{N_s} $, where $ N_s $ denotes the number of samples sampled from a distribution $ p $. $ y_i \in \mathcal{Y} $ represents the ground-truth corresponding to the image $ x_i $. Besides, the wireless channel condition is denoted as $ \bm{\sigma}^ \mathcal{S} $. We aim to deploy our model into a new environment referred to as the target domain with $N_t$ unlabeled data $\mathcal{D}^{\mathcal{T}} = (x_j^{{\mathcal{T}}_1}, \cdots, x_j^{{\mathcal{T}}_K})_{j=1}^{N_t}$ sampled from a distribution $ q $ and has wireless channel condition $ \bm{\sigma} ^ \mathcal{T} $. \par
In practice, the target domain may differ significantly from the source domain in both data and channel distributions due to the different deployment locations. Let us denote $ \theta^\mathcal{S} $ and $ \theta^\mathcal{T} $ as the DNN model in the source domain and target domain, respectively. With abundant labeled data in the source domain, we can apply conventional supervised learning to optimize the DNN model $ \theta^{\mathcal{S}} $ to attain satisfactory inference accuracy. As we assume the target domain has no labeled data, conventional supervised learning is inapplicable to train $ \theta^{\mathcal{T}} $. Meanwhile, as we will show in Section \ref{IV}, direct deployment of $ \theta^{\mathcal{S}} $ to the target domain may result in significant performance degradation. In the next section, we propose an efficient transferable deployment method DASEIN to maximize the edge inference accuracy in the target domain. 

\section{The Proposed DASEIN Method}
\label{III}

\begin{figure}[t]%调节图片位置，h：浮动；t：顶部；b:底部；p：当前位置
	\centering
	\includegraphics[width=5.5cm]{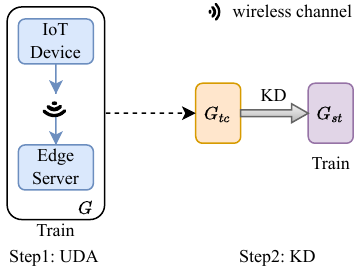}  % 中括号中的为调节图片大小
	\caption{The simplified flowchart of DASEIN system. Step 1 mitigates the variations in data distribution by training model $ G $ through UDA. Step 2 further mitigates the simultaneous variations of channel distributions by utilizing G as the teacher model $ G_{tc} $ to train a new student model $ G_{st} $.}
	\label{simplified flowchart}%文中引用该图片代号
\end{figure}

In this section, we describe the methodology of the proposed DASEIN method for transferable deployment of edge inference system. As shown in Fig. \ref{simplified flowchart}, the training process of the DASEIN model consists of two steps. In step 1, model $ G $ is trained under favorable channel conditions using UDA to align the distribution discrepancy between unlabeled and labeled data. In step 2, the trained $ G $ serves as a teacher model $ G_{tc} $, guiding the student model $ G_{st} $ through KD to facilitate adaptation to its specific wireless channel condition.

%As shown in Fig. \ref{simplified flowchart}, the training of DASEIN consists of two steps that sequentially adapt to the data and channel distributions, using UDA and KD techniques, respectively. 

\subsection{Methodology\label{III-A}}
\subsubsection{Unsupervised Domain Adaptation\label{III-A1}}
\begin{figure*}[t]
	\centering
	\includegraphics[width=11cm]{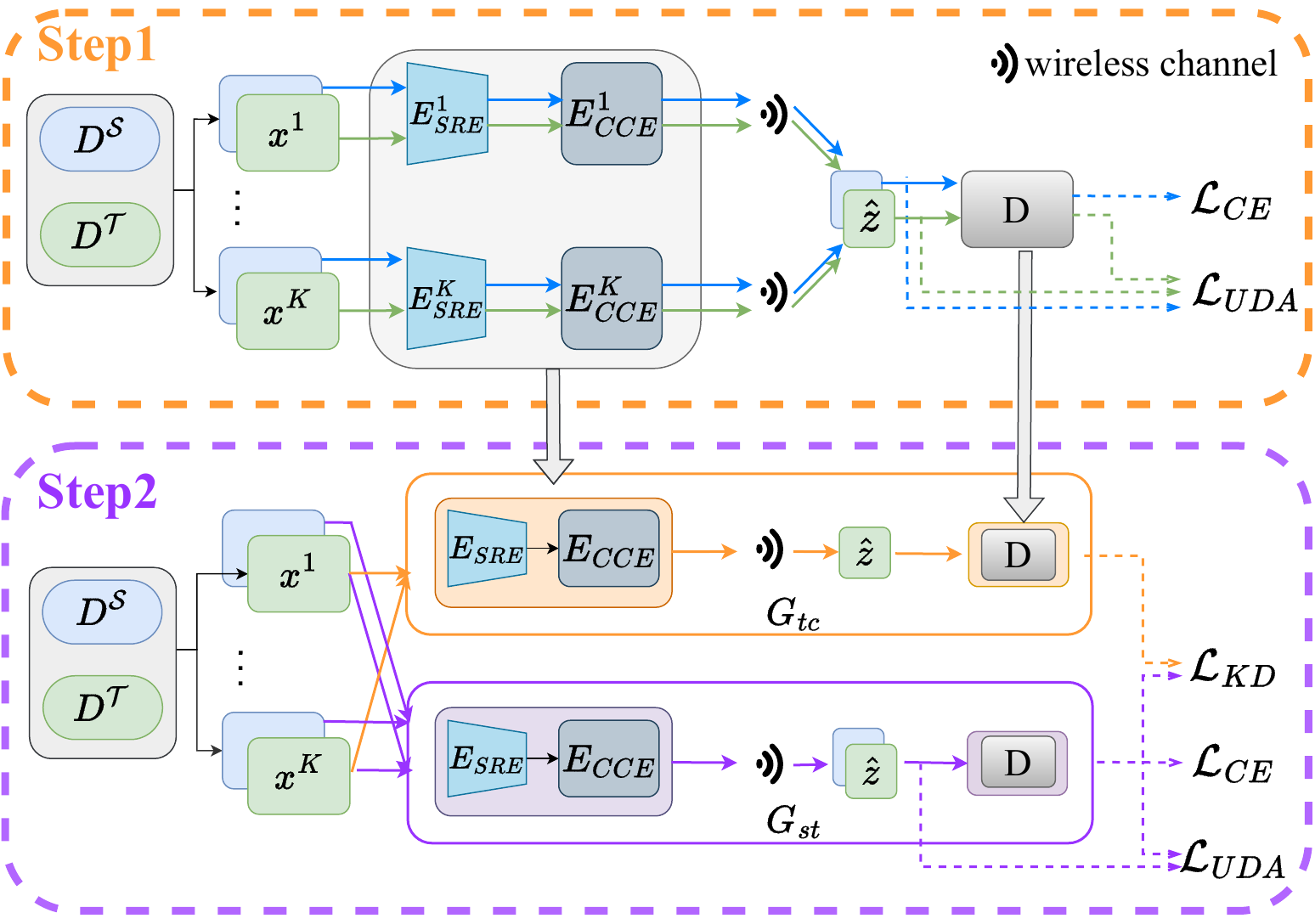}
	\caption{Step 1 mitigates the variations in data distribution by minimizing cross-entropy loss $ \mathcal{L}_{CE} $ and UDA loss $ \mathcal{L}_{UDA} $, whereas step 2 further mitigates the simultaneous variations of channel distributions by minimizing $ \mathcal{L}_{CE} $, $ \mathcal{L}_{UDA} $, and knowledge distillation loss $ \mathcal{L}_{KD} $.}
	\label{system diagram}
\end{figure*}

At this stage, we assume for the moment that the SNR in the source and target domains are identical and focus on adapting to the new data distribution. As shown in Fig. \ref{system diagram}, DASEIN inputs both the source domain data $ \mathcal{D}^{\mathcal{S}} $ and target domain data $ \mathcal{D}^{\mathcal{T}} $ to the corresponding $ E_{SRE} $ and $ E_{CCE} $ of the $ K $ target domain sensing devices, which generate two sets of semantic encoding sequences $ z^{\mathcal{S}_k} $ and $ z^{\mathcal{T}_k} $, where $ k = 1, 2, \cdots, K $. The edge server combines the noisy $ \hat{z}^{\mathcal{S}} $ and $ \hat{z}^{\mathcal{T}} $ through the wireless channel. The goal of UDA is to align the features of the source and target domain data in a latent space. \par
To measure the discrepancy of data distributions, we first consider a domain adaptation metric named maximum mean discrepancy (MMD) \cite{a29}. Specifically, it measures the discrepancy between two distributions $ p $ and $ q $ by computing 
\begin{equation}\label{e5}
	d(p,q) = \left\| E_p[\phi(x^{\mathcal{S}})] - E_q[\phi(x^{\mathcal{T}})] \right\| ^2_{\mathcal{H}},
\end{equation}
where $ \mathcal{H} $ is the Reproducing Kernel Hilbert Space (RKHS) endowed with a characteristic kernel $ \mathcal{K} $, the function $ \phi(\cdot) $ (detailed in Appendix \ref{A}) maps the data samples to the RKHS. In practice, we use the observed samples to compute the following unbiased estimation of \eqref{e5}:
\begin{equation}\label{e6}
	\begin{aligned}
		\hat{d}(p,q) &= \left\| \frac{1}{n_s} \sum_{x_i \in \mathcal{D}^\mathcal{S}} \phi(x_i) - \frac{1}{n_t} \sum_{x_j \in \mathcal{D}^\mathcal{T}} \phi(x_j) \right\| ^2_{\mathcal{H}} \\
		&=\frac{1}{n_s^2} \sum_{i=1}^{n_s} \sum_{j=1}^{n_s} \mathcal{K}(x_i^{\mathcal{S}}, x_j^{\mathcal{S}}) + \frac{1}{n_t^2} \sum_{i=1}^{n_t} \sum_{j=1}^{n_t} \mathcal{K}(x_i^{\mathcal{T}}, x_j^{\mathcal{T}}) \\
		&- \frac{2}{n_s n_t} \sum_{i=1}^{n_s} \sum_{j=1}^{n_t} \mathcal{K}(x_i^{\mathcal{S}}, x_j^{\mathcal{T}}),
	\end{aligned}
\end{equation}
where we apply the Gaussian kernel function $ \mathcal{K}(x^{\mathcal{S}}, x^{\mathcal{T}}) = exp(- \Vert x^{\mathcal{S}} - x^{\mathcal{T}} \Vert^2 / 2 \sigma_b ^2 ) $ to compute the mapping $ \phi(\cdot) $, with $ \sigma_b $ being a bandwidth parameter controlling the range of $ \mathcal{K}(\cdot,\cdot) $. In the RKHS, inner products can be computed using kernels as $ \langle \phi(x^\mathcal{S}), \phi(x^\mathcal{T}) \rangle = \mathcal{K}(x^\mathcal{S}, x^\mathcal{T}) $, where we show the detailed computations in Appendix \ref{A}.

Notice that crudely minimizing the MMD of two data distributions may lose fine-grained categorical information. For this, we introduce the LMMD \cite{a25} to calculate the category-weighted discrepancies between compressed semantic features $ \hat{z}^{\mathcal{S}} $ and $ \hat{z}^{\mathcal{T}} $ extracted from the observation datasets $ \mathcal{D}^\mathcal{S} $ and $ \mathcal{D}^\mathcal{T} $, i.e.,
\begin{equation}\label{e7}
	\begin{aligned}
		&\hat{d}(p,q) = \frac{1}{C}\sum_{c=1}^C \left\| \sum_{x_i^{\mathcal{S}} \in \mathcal{D}^{\mathcal{S}}} \omega_i^{{\mathcal{S}}c} \phi(\hat{z}_i^{\mathcal{S}}) - \sum_{x_j^{\mathcal{T}} \in \mathcal{D}^{\mathcal{T}}} \omega_j^{{\mathcal{T}}c} \phi(\hat{z}_j^{\mathcal{T}}) \right\|^2_{\mathcal{H}} \\
		&= \frac{1}{C}\sum_{c=1}^C \biggl[ \sum_{i=1}^{n_s}\sum_{j=1}^{n_s} \omega_i^{{\mathcal{S}}c} \omega_j^{{\mathcal{S}}c} \mathcal{K}(\hat{z}_i^{\mathcal{S}},\hat{z}_j^{\mathcal{S}}) \\ 
		&+ \sum_{i=1}^{n_t}\sum_{j=1}^{n_t} \omega_i^{{\mathcal{T}}c} \omega_j^{{\mathcal{T}}c} \mathcal{K}(\hat{z}_i^{\mathcal{T}},\hat{z}_j^{\mathcal{T}}) - 2\sum_{i=1}^{n_s}\sum_{j=1}^{n_t} \omega_i^{{\mathcal{S}}c} \omega_j^{{\mathcal{T}}c} \mathcal{K}(\hat{z}_i^{\mathcal{S}},\hat{z}_j^{\mathcal{T}})\biggr].
	\end{aligned}
\end{equation}
Here, $ \omega_i^{*c} $ represents the weight of the class c, i.e.,
\begin{equation}\label{e_2}
	\omega_i^{*c} = y_{ic}^*/\sum\nolimits_{(x_j,y_j) \in \mathcal{D}^*} y_{jc}^*,
\end{equation}
where $ * $ indicates domain $ \mathcal{S} $ or $ \mathcal{T} $, $ y_{ic}^* $ denotes the $ c $-th binary entry in the one-hot vector $ y_{oh, i}^* $ of label $ y_{i}^* $. While $ w_i^{\mathcal{S}c} $ can be directly calculated from source domain ground-truths $ y^\mathcal{S} $, $ w_j^{\mathcal{T}_c} $ is calculated from predicted labels $ \hat{y}^{\mathcal{T}} $ due to the lack of ground truth. As described in Section \ref{II-A}, $ \hat{y}^{\mathcal{T}} $ is the predicted label of the target domain data $\mathcal{D}^{\mathcal{T}}$ at the edge server. The result computed via \eqref{e7} is referred to as the \textit{domain adaptation loss}, which will be minimized to align the source and target distributions. The detailed training procedure will be given in Section \ref{III-C}. With a bit abuse of notation, we denote the target domain DNN models after training convergence as $\theta^{\mathcal{T}}$. 

\subsubsection{Knowledge Distillation\label{III-A2}}
%KD aims to train the student model $ G_{st} $ to match the teacher model $ G_{tc} $, resulting in a lightweight model with similar performance\cite{a29-1}. 
KD aims to train the student model $ G_{st} $ to match the teacher model $ G_{tc} $, resulting in a lightweight model with similar performance \cite{a29-1}, and it has been widely applied across various domains \cite{a29-2}. $ G_{tc} $ is employed to generate soft labels $ \bm{\hat{p}}_{tc} $ for training dataset to guide $ G_{st} $, which is a probability distribution. KD not only minimizes the distance $ d\left( \bm{\hat{p}}_{st}, \bm{\hat{p}}_{tc} \right) $ between predicted probability distribution $ \bm{\hat{p}}_{st} $ of $ G_{st} $ and the soft label $ \bm{\hat{p}}_{tc} $, but also minimizes the distance $ d\left( y, \bm{\hat{p}}_{st} \right) $ between $ \bm{\hat{p}}_{st} $ and its ground-truth $ y $ for training dataset, i.e.,
	\begin{equation}\label{e_distillation}
		d = \alpha d\left( \bm{\hat{p}}_{st}, \bm{\hat{p}}_{tc} \right) + \beta d\left( y, \bm{\hat{p}}_{st} \right),
	\end{equation}
	where $ \alpha $ and $ \beta $ are hyperparameters for controlling the relative importance of the two parts of the loss. \par
After aligning the source and target domain data distributions in step 1, we continue to address the discrepancy in the channel condition by \textit{knowledge distillation}. Because changing the channel condition $ \bm{\sigma} $ does not vary the labels of data samples, we allow the $ G_{tc} $ to obtain the predicted result $ \bm{\hat{p}}^{\mathcal{T}}_{tc} $ for the $ \mathcal{D}^{\mathcal{T}} $ to guide the $ G_{st} $. As shown in step 2 of Fig. \ref{system diagram}, we introduce two models, i.e., a teacher model with parameter $ \theta_{tc} = \{ \theta_{SRE}^{tc}, \theta_{CCE}^{tc}, \theta_{D}^{tc} \} $ and a student model $ \theta_{st} = \{ \theta_{SRE}^{st}, \theta_{CCE}^{st}, \theta_{D}^{st} \} $. Initially, we set $ \theta_{tc} = \theta_{st} = \theta^{\mathcal{T}} $. \par
We initially calculate the difference $ d\left( \bm{\hat{p}}^{\mathcal{T}}_{st} , \bm{\hat{p}}^{\mathcal{T}}_{tc} \right) $ between the predicted categorical probability distributions $\bm{\hat{p}}^{\mathcal{T}}_{tc}$ from $G_{tc}$ and $\bm{\hat{p}}^{\mathcal{T}}_{st}$ from $G_{st}$ on the $\mathcal{D}^{\mathcal{T}}$ in \eqref{e_distillation}. To enhance the robustness of KD, we utilize the confidence threshold as an evaluation criterion for $ \bm{\hat{p}}^{\mathcal{T}}_{tc} $. $ \bm{\hat{p}}^{\mathcal{T}}_{tc,max} $ is the maximum entry in $ \bm{\hat{p}}^{\mathcal{T}}_{tc} $. A higher $ \bm{\hat{p}}^{\mathcal{T}}_{tc,max} $ indicates a higher probability of accurately predicted result $ \bm{\hat{p}}^{\mathcal{T}}_{tc} $. We filter unreliable predictions of data samples by implementing a binary mask $ m $. When $ \bm{\hat{p}}^{\mathcal{T}}_{tc,max,i} $ of a sample $ x_i^{\mathcal{T}} $ exceeds a certain confidence threshold $ \epsilon $, we consider the result reliable and set the mask $ m=1 $. Conversely, when $ \bm{\hat{p}}^{\mathcal{T}}_{tc,max,i} $ falls below $ \epsilon $, we exclude this sample from fine-tuning in this round and set the mask for $ x_i^{\mathcal{T}} $ as $ m=0 $. That is, 
	\begin{equation} \label{e10}
		m(x_i^{\mathcal{T}})=\left\{
		\begin{aligned}
			1 & , & \text{if} \ \bm{\hat{p}}^{\mathcal{T}}_{tc,max,i}>\epsilon, \\
			0 & , & otherwise.
		\end{aligned}
		\right.
	\end{equation}
	We preserve only the data samples with the mask equal to 1 and denote the predicted results of $ G_{tc} $ and $ G_{st} $ as $ \bm{\hat{p}}_{tc, m}^{\mathcal{T}} $ and $ \bm{\hat{p}}_{st, m}^{\mathcal{T}} $. The optimization of $ G_{st} $ is related to the calculation of the cross-entropy between $ \bm{\hat{p}}_{tc, m}^{\mathcal{T}} $ and $ \bm{\hat{p}}_{st, m}^{\mathcal{T}} $, i.e., 
	\begin{equation}\label{e_1}
		d(\bm{\hat{p}}^{\mathcal{T}}_{st, m} , \bm{\hat{p}}^{\mathcal{T}}_{tc, m}) = -\sum_{c=1}^{C} \bm{\hat{p}}^{\mathcal{T}}_{st, m, c} \log(\bm{\hat{p}}^{\mathcal{T}}_{tc, m, c}).
\end{equation} \par
Due to the lack of ground-truths in the target domain $\mathcal{T}$, computing $ d\left( y, \bm{\hat{p}}_{st} \right) $ for $\mathcal{D}^{\mathcal{T}}$ in Equation \eqref{e_distillation} is impossible. To address this issue, we employ labeled source domain data $\mathcal{D}^{\mathcal{S}}$ to calculate $ d(y_{oh}^{\mathcal{S}}, \bm{\hat{p}}^{\mathcal{S}}_{st}) $ between the predicted probability distribution $ \bm{\hat{p}}^{\mathcal{S}}_{st} $ from $ G_{st} $ for $\mathcal{D}^{\mathcal{S}}$ and the one-hot vector ${y}_{oh}^{\mathcal{S}}$ of ground-truths $ y^{\mathcal{S}} $:
	\begin{equation}\label{e_3}
		d(y_{oh}^{\mathcal{S}}, \bm{\hat{p}}^{\mathcal{S}}_{st}) = -\sum_{c=1}^{C} y_{oh, c}^{\mathcal{S}} \log(\bm{\hat{p}}^{\mathcal{S}}_{st, c}).	
	\end{equation}
The idea of using $ \mathcal{D}^{\mathcal{S}} $ is to leverage the classification performance of $ G_{st} $ on the source dataset to improve its classification performance on $ \mathcal{D}^{\mathcal{T}} $. Hence, the UDA loss term in \eqref{e7} is added in KD, the loss used for gradient updates during UDA, to ensure that the classification performance remains consistent across both domains.
% Hence, the UDA loss term in \eqref{e7} is added in KD to ensure that the classification performance remains consistent across both domains.

	\subsection{Network Design\label{III-B}}
In DASEIN, the target domain model $ \theta^{\mathcal{T}} $, teacher model $ \theta_{tc} $, and student model $ \theta_{st} $ use the same DNN structure as following:
\subsubsection{Semantic Representation Extractor\label{III-B1}}
We implement a ResNet \cite{a30} network by removing its output classifier as $ E_{SRE} $, which leverages the advantage of the powerful feature extraction capabilities of ResNet. 
\subsubsection{Compress and Channel Encoder\label{III-B2}}
We use a fully connected (FC) layer as the $ E_{CCE} $ and define the compression rate (CR) as $ a_{out} / a_{in} $. A lower CR removes more redundancy, however, may also lose critical semantic information and decrease the overall inference accuracy, and vice versa.
\subsubsection{Decoder\label{III-B3}}
The output of the decoder is the inference result of the task. We build the decoder using linear and rectified linear unit (ReLU) layers, where the linear layer performs a weighted transformation on the input, and the ReLU layer introduces non-linearity, allowing the network to capture complex information. \par

\subsection{Training Strategy\label{III-C}}

The training consists of two steps, which correspond to the UDA and KD procedures, respectively.
\subsubsection{Step 1\label{III-C1}}
First, we compute the classification loss using only the source domain dataset $ \mathcal{D}^{\mathcal{S}} $. The loss function can be constructed by the predictions $ \bm{\hat{p}}^\mathcal{S} $ in \eqref{e4} and the ground-truth $ y^\mathcal{S}_{oh} $ of the sample $ x^\mathcal{S} $:
\begin{equation}\label{e8}
	\mathcal{L}_{CE} = -\sum_{c=1}^{C} y_{oh, c}^{\mathcal{S}} \log(\bm{\hat{p}}^{\mathcal{S}}_{c}),	
\end{equation} \par

Next, using both the source and target domain data, we denote $ \mathcal{L}_{UDA} = \hat{d}(p,q) $ as the LMMD loss computed from \eqref{e7}, and the total training loss is:
\begin{equation}\label{e9}
	\mathcal{L}_1 = \mathcal{L}_{CE} + \delta \lambda \mathcal{L}_{UDA},
\end{equation}
where $ \lambda > 0 $ is a weighting parameter. Here, $ \delta $ can be expressed as: 
	\begin{equation}\label{e_4}
		\delta = \frac{2}{1 + \exp\left(-10 \cdot \frac{\text{e}}{E}\right)} - 1,
	\end{equation}
	where $ e $ refers to the current training epoch, $ E $ denotes the total epoch. The increase of $ \delta $ with the $ e $ serves as a ``warm-up" period that gradually enhances the importance of UDA loss. As outlined in Section \ref{III-A1}, the LMMD loss relies on the predicted probability distribution $ \bm{\hat{p}}^{\mathcal{T}} $ of model. During the initial stage of training, these predictions are often inaccurate, which may misguide the domain adaptation process. To address this issue, $ \delta $ initially is very small such that the model primarily performs supervised learning based on $ \mathcal{D}^\mathcal{S} $. As the training progresses, the model gains the ability to accurately classify $ \mathcal{D}^\mathcal{S} $ and can utilize domain invariance to classify $ \mathcal{D}^\mathcal{T} $. Increasing $ \delta $ as training rounds allows the model to gradually adapt to the target domain. The detailed pseudo-code of training step 1 is given in Algorithm \ref{alg:a1}. \par

\subsubsection{Step 2\label{III-C2}}

\begin{algorithm}[!t]
	\caption{Training Step 1 of DASEIN}
	\label{alg:a1}  
	\begin{algorithmic}[1]
		\State \textbf{Input:} $ \mathcal{D}^{\mathcal{S}} $; $ \mathcal{D}^{\mathcal{T}} $; batchsize $ n_b $; epoch $ E $.
		\State \textbf{Output:} optimized network parameters $ \theta $.
		\State initialize $ \theta $.
		\For{$e = 1, 2, \cdots $ to $E$} 
		\State get $ ((x_i^{{\mathcal{S}}_k}, y_i^{{\mathcal{S}}_k})_{i=1}^{n_b}, (x_j^{{\mathcal{T}}_k})_{j=1}^{n_b}) $ in ($ \mathcal{D}^{\mathcal{S}} $, $  \mathcal{D}^{\mathcal{T}} $). 
		\State compute loss function $ \mathcal{L}_1 $ in \eqref{e9}.  
		\State $ \theta \leftarrow \theta - \eta\bigtriangledown\mathcal{L}_1 $, $ e \leftarrow e + 1 $.
		\EndFor 
	\end{algorithmic}
\end{algorithm}

\begin{algorithm}[!t]
	\caption{Training Step 2 of DASEIN}
	\label{alg:a2}  
	\begin{algorithmic}[1]
		\State \textbf{Input:} $ \mathcal{D}^{\mathcal{S}} $; $ \mathcal{D}^{\mathcal{T}} $; $ \theta_{tc} $; batchsize $ n_b $; finetune epoch $ E_f $.
		\State \textbf{Output:} optimized student model parameters $ \theta_{st} $.
		\State initial $ \theta_{st} = \theta_{tc} = \theta^{\mathcal{T}} $.
		\For{$e = 1, 2, \cdots $ to $ E_f $} 
		\State get $ ((x_i^{{\mathcal{S}}_k}, y_i^{{\mathcal{S}}_k})_{i=1}^{n_b}, (x_j^{{\mathcal{T}}_k})_{j=1}^{n_b}) $ in ($ \mathcal{D}^{\mathcal{S}} $, $  \mathcal{D}^{\mathcal{T}} $). 
		\State compute loss function $ \mathcal{L}_2 $ in \eqref{e12}.  
		\State $ \theta_{st} \leftarrow \theta_{st} - \eta\bigtriangledown\mathcal{L}_2 $, $ e \leftarrow e + 1 $
		\EndFor
	\end{algorithmic}
\end{algorithm}

As shown in Fig. \ref{system diagram}, the fine-tuning process requires both source domain $ \mathcal{D}^{\mathcal{S}} $ and target domain data $\mathcal{D}^{\mathcal{T}} $, which is simultaneously fed into the teacher and student models. As discussed in Section \ref{III-A2}, the loss function contains $ d(y_{oh}^{\mathcal{S}}, \bm{\hat{p}}^{\mathcal{S}}_{st}) $ of source domain data $ \mathcal{D}^{\mathcal{S}} $ cross-entropy in \eqref{e_3}, the UDA loss $ \hat{d}(p,q) $ in \eqref{e7} of target domain data $ \mathcal{D}^{\mathcal{T}} $, and $ d(\bm{\hat{p}}^{\mathcal{T}}_{st, m} , \bm{\hat{p}}^{\mathcal{T}}_{tc, m}) $ in \eqref{e_1} during the fine-tuning process, which is expressed as:
	\begin{equation}\label{e12}
		\mathcal{L}_2 = d(y_{oh}^{\mathcal{S}}, \bm{\hat{p}}^{\mathcal{S}}_{st}) + \lambda_1 \hat{d}(p,q) + \lambda_2 d(\bm{\hat{p}}^{\mathcal{T}}_{st, m} , \bm{\hat{p}}^{\mathcal{T}}_{tc, m}),
	\end{equation}
	where $ \lambda_1 > 0 $ and $ \lambda_2 > 0 $ are the weighting coefficients. Algorithm \ref{alg:a2} provides the pseudo-code of training step 2.\par

Compared to conventional KD that uses ground-truth hard labels to train the student model in a supervised-learning fashion, in the absence of ground-truth labels in the target domain, we consider both soft and hard labels in \eqref{e12} to guide the knowledge transfer to student model, and adopt a sampled filtering mechanism to ensure the reliability of soft labels.

\section{Digital Implementation of DASEIN}\label{III2}

\subsection{Digital Semantic Transceivers}\label{III2-A}
\begin{figure}[t]%调节图片位置，h：浮动；t：顶部；b:底部；p：当前位置
	\centering
	\includegraphics[width=9cm]{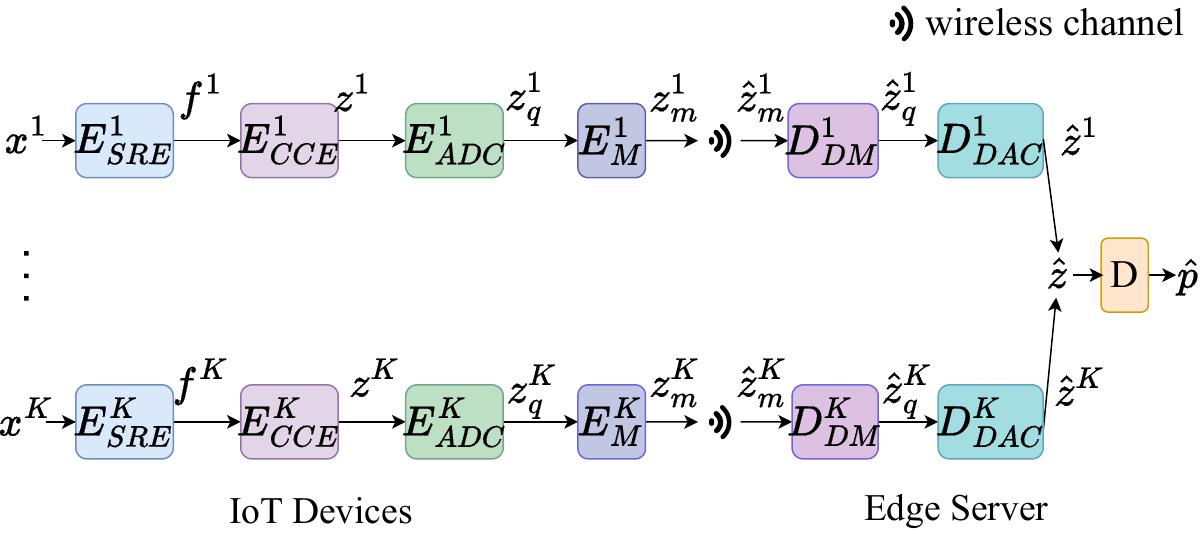}% 中括号中的为调节图片大小
	\caption{The digital schematics of the considered edge inference system.}
	\label{4digital}%文中引用该图片代号
\end{figure}

The proposed edge sensing system relies on analog transmission of continuous features. In practical implementation, it is susceptible to noise, interference and signal attenuation during transmission. Besides, it is not compatible with modern communication systems based on digital modulation techniques. For this, we consider a digital implementation of the proposed DASEIN scheme shown in Fig. \ref{4digital}. The $ k $-th IoT device converts the extracted feature $ z^k $ to a digital signal $ z_q^k $ via an Analog-to-Digital Converter (ADC) $ E_{ADC}^k(\cdot) $. We denote the minimum and maximum of $ z^k, \forall k $, as $ z_\text{min} $ and $ z_\text{max} $, respectively. In practice, we can introduce an bounded function, e.g., tanh($ \cdot $), after the output layer of $ E_{CCE}^k $ to ensure that $ z^k\in(z_{min}, z_{max}) $. Let $ q_b $ be the quantization resolution. The ADC uniformly quantizes the signal range $ z_\text{max} - z_\text{min} $ into $ 2^{q_b} $ levels and maps $ z^k $ into an interger index $ z_{id}^k $ within $[0, 2^{q_b}-1]$. The value of $ z_{id}^k $ is computed as:
	\begin{equation}\label{e_10_11}
		z_{id}^k = \text{round} \left( g(z^k) \right) , k = 1,2,\cdots, K,
	\end{equation}
	where round($ \cdot $) denotes the nearest interger function and
	\begin{equation}\label{e_0_11}
		g(z^k) = \frac{z^k - z_\text{min}}{z_\text{max} - z_\text{min}} \left( 2^{q_{b}} - 1 \right) , k = 1,2,\cdots, K.
	\end{equation}
	The ADC converts $ z_{id}^k $ into a binary code $ z_q^k $ using natural binary coding and feed it into a digital modulator $ E_M^k(\cdot) $ to obtain the modulated symbols $ z^k_m $ for wireless transmission (e.g., quadrature phase shift keying (QPSK) symbols under $2$-bit quantization), i.e.,
	\begin{equation}\label{e_6}
		z^k_m = E_M^k (z^k_q), k = 1, 2, \cdots, K.
	\end{equation}
	At the edge server, the received symbol is
	\begin{equation}\label{e_6_1}
		\hat{z}_m^k = z^k_m + n^k, k = 1,2,\cdots, K.
	\end{equation}
	We first employ a digital demodulator $ D_{DM}(\cdot) $ to recover the binary code from $ \hat{z}_m^k $ as
	\begin{equation}\label{e_7}
		\hat{z}^k_q = D_{DM}( \hat{z}^k_m ), k = 1, 2, \cdots, K,
	\end{equation}
	which is then processed by a Digital-to-Analog Convertor (DAC) $ D_{DAC}(\cdot) $ to restore the analog featured
	\begin{equation}\label{e_11}
		\hat{z}^k = g^{-1} \left( \hat{z}_{id}^k \right), k = 1,2,\cdots, K,
	\end{equation}
where $ \hat{z}_{id}^k $ is the decimal conversion of $ \hat{z}_q^k $, $ g^{-1}(\cdot) $ is the inverse of $ g(\cdot) $ in \eqref{e_0_11}. Following the same procedures in Section \ref{II-A2}, we use $ \hat{z}_k $ for the subsequent task inference.

\subsection{Training Method}\label{III2-B}
The non-differentiable round process in \eqref{e_10_11} poses a great challenge to the gradient-based AI model training process. To address this problem, we apply a surrogate quantization function \cite{a30_1} to approximate the round($ \cdot $) function:
	\begin{equation} \label{e_9}
		R_{\sin}(x, r)=\left\{
		\begin{aligned}
			&\frac{x - \sin(2\pi x)}{2\pi}, &\text{if} \  r=1, \\
			&R_{\sin}(R_{\sin}(x, r-1), 1), &\text{if} \  r>1. 
		\end{aligned}
		\right.
	\end{equation}
Here, $ x \in \mathcal{R} $ is a real-valued input and $ r \ge 1 $ is a positive integer parameter that controls the precision of approximation. As shown in \cite{a30_1}, $ R_{\sin}(x, r) $ provides a differentiable and close approximation to torch.round() function when $r=3$. \par
During the training phase, we bypass the modulation and demodulation procedures and transmit the output of $ R_{sin} $, denoted as $ \tilde{z}_{id}^k $ (i.e., an approximation of $ z_{id}^k $), to the edge server. Notice that complex noise $ n^k $ is considered in \eqref{e_6_1} during the transmission of modulated signal in inference phase. For consistency, we consider a real-valued noise $ n_{real}^k $ with the same power to $ n^k $ for training. Correspondingly, the received signal at the edge server is 
	\begin{equation}\label{e_11_1}
		\hat{\tilde{z}}_{id}^k = \tilde{z}_{id}^k + n_{real}^k, k = 1,2,\cdots, K.
	\end{equation}
It is then fed into the DAC to restore the analog feature, and subsequently passed to the decoder to obtain the prediction $ \hat{p} $ in \eqref{e4}. Finally, we update the model parameters following the same training procedures in Section \ref{III}.

\section{Performance Evaluation} \label{IV}
\subsection{Experiment Setup\label{IV-A}}
\subsubsection{Dataset\label{IV-A1}} 
\begin{figure}[t]
	\centering
	\subfloat[The same class of item in datasets L, S, and V.]{\includegraphics[width=0.81\hsize, height=0.27\hsize]{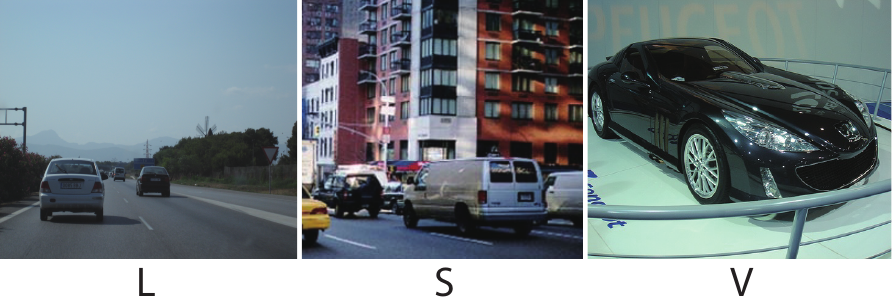}\label{datasetVLS}}
	\hfill
	\subfloat[Data splitting for multi-view observations.]{\includegraphics[width=0.81\hsize, height=0.27\hsize]{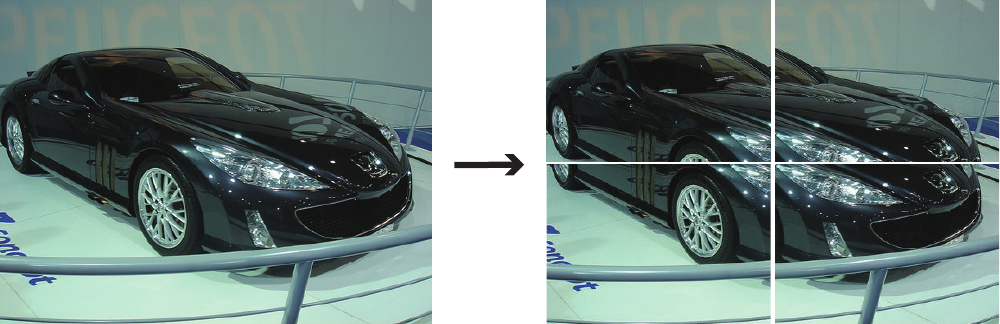}\label{seperate}}
	\caption{The description of the OFFICE-31 dataset.}
	\label{dataset}
\end{figure}

OFFICE-31 dataset \cite{a31} includes images of 31 common office items. Each item has three different image styles: the Amazon image dataset (A) with 2,817 images, the low-resolution image dataset captured by webcams (W) with 795 images, and the high-resolution image dataset captured by DSLR cameras (D) with 498 images. The W and D datasets depict the same office items but differ in resolution. The images of A have different item styles and backgrounds compared to those in W and D. For example, we denote a transfer task from source domain A to target domain W as A→W. VLCS dataset \cite{a32} consists 5 distinct categories: bird, car, chair, dog, and person. We utilize three different image styles from the VLCS dataset: LabelMe (L) contains 2,656 images, SUN09 (S) consists of 3,282 images, and VOC2007 (V) includes 3,376 images. Similarly, we denote a transfer task from source domain L to target domain S as L→S. An illustrative example is shown in Fig. \ref{datasetVLS}. The majority of "car" samples in L are from suburban streets, while in S, they predominantly come from urban streets. The samples predominantly consist of close-up images of “car” in V. We employ these three sets of image datasets to simulate the change of data distribution in different deployment scenarios.
\subsubsection{Model Details\label{IV-A2}}
Table \ref{tab:1} shows the detailed structure of the DNN model. We utilize a ResNet50 network without its FC layer-based classifier for feature extraction whose output dimension is 2048. The compressor output dimension can then be set according to a prescribed CR. Without loss of generality, we normalize the output of the semantic compressor to limit the transmit power to 1. To enhance the classification accuracy at the edge server, we use two linear layers as the classifier. \par

\begin{table}[t]
	\centering
	\caption{The Details Structure of the System Model}
	\label{tab:1}
	\renewcommand{\arraystretch}{1.5}
	\begin{tabular}{c c c} % 让第一列自动换行
		\toprule
		\textbf{Component} & \textbf{Layer} & \textbf{Output Dimensions} \\
		\midrule
		Input & image & $150 \times 150$ \\
		\midrule
		\multirow{2}{*}{\makecell[c]{Semantic Representation \\ Extractor}} & ResNet50 & \multirow{2}{*}{2048} \\
		& (without classifier) & \\
		\midrule
		\multirow{2}{*}{\makecell[c]{Compress and \\ channel encoder}} & ReLU & 2048 \\
		& Linear & $a_{out}$ \\
		\midrule
		Channel & AWGN & $a_{out}$ \\
		\midrule
		Concatenate & torch.cat & $4 \times a_{out}$ \\
		\midrule
		\multirow{4}{*}{Decoder} & Linear & 256 \\
		& ReLU & 256 \\
		& Linear & 31 \\
		& Softmax & 31 \\
		\bottomrule
	\end{tabular}
\end{table}

\subsubsection{Experiment Details\label{IV-A3}}

\begin{table*}[ht]
	\centering
	\caption{Training parameters of DASEIN}
	\label{tab:parameter} 
	\renewcommand{\arraystretch}{2.0} % 调整行高，1.5 是行高的倍数
	\setlength{\tabcolsep}{6.0pt} % 设置列之间的间隔
	\begin{tabular}{cccccccccccc}
		\toprule
		Batch size & $ E $ & $ {\eta_0}\left( E_{SRE}\right)  $ & $ {\eta_0}\left( E_{CCE}\right)  $ & $ {\eta_0}\left( D\right) $ & Optimizer & Momentum & Weight decay & $ \lambda $ & $ E_f $ & $ \lambda_1 $ & $ \lambda_2 $ \\
		\hline
		16 & 100 & $10^{-3}$ & $10^{-2}$ & $10^{-2}$ & SGD & 0.9 & $5 * 10^{-4}$ & 0.1 & 20 & 0.1 & 0.5 \\
		\hline
	\end{tabular}
\end{table*}

All the images are cropped into 224 × 224 three-channel pixels. As illustrated in Fig. \ref{seperate}, to simulate the multi-view image captured by different IoT cameras, we divide an image into four parts, which are overlapped with 150 × 150 pixels. Table \ref{tab:parameter} presents the main training parameters. In step 1, model updates are performed using the stochastic gradient descent (SGD) with a momentum of 0.9. The learning rate of the pre-trained Resnet50 network is set to 0.001. Besides, the model learning rate of the $ E_{CCE} $ and Decoder is 0.01. For all tasks, we employ a learning rate annealing strategy: learning rate $ \eta = \eta_0 / (1 + 10 (e/E))^{0.75} $, where $ \eta_0 $ is the initial learning rate, $ e $ refers to the current training epoch, $ E $ denotes the total epoch. The values of $ E $ is 100, $ \lambda $ is equal to 0.1. In step 2, $ E_f $ is equal to 20, $ \lambda_1 $ and $ \lambda_2 $ are respectively equal to 0.1 and 0.5, with all other parameters remaining the same as in step 1. All the computations are executed on a machine with an Intel(R) Xeon(R) Gold 6142M CPU, a NVIDIA Tesla P40 GPU, and 94.7GB RAM. All the proposed algorithms are written and evaluated using Python 3.7, which is compatible with popular machine learning libraries. The adopted hardware, software, and toolbox suffice to conduct the considered multimedia processing and AI model training/inference tasks. For performance comparisons with the proposed DASEIN method, we consider the following three benchmarks:
%  这个是之前的版本，保留一下：
%  All the images are cropped into 224 × 224 three-channel pixels. As illustrated in Fig. \ref{seperate}, to simulate the multi-view image captured by different IoT cameras, we divide an image into four parts, which are overlapped with 150 × 150 pixels. Model updates are performed using the stochastic gradient descent (SGD) with a momentum of 0.9. The learning rate (L) of the pre-trained Resnet50 network is set to 0.001. Besides, the model learning rate of the CCE and encoder is 0.01. For all tasks, we employ a learning rate annealing strategy: learning rate $ \eta = \eta_0 / (1 + 10 (e/E))^{0.75} $, where $ \eta_0 $ is the initial learning rate, $ e $ refers to the current training epoch, $ E $ denotes the total epoch. The values of $ E $ and $ E_f $ are 100 and 20, respectively. $ \lambda $ is equal to 0.1, $ \lambda_1 $ and $ \lambda_2 $ are respectively equal to 0.1 and 0.5. The confidence threshold $ \epsilon $ is set to 0.6. All the computations are executed on a machine with an Intel(R) Xeon(R) Gold 6142M CPU, a NVIDIA Tesla P40 GPU, and 94.7GB RAM. All the proposed algorithms are written and evaluated using Python 3.7, which is compatible with popular machine learning libraries. The adopted hardware, software, and toolbox suffice to conduct the considered multimedia processing and AI model training/inference tasks. For performance comparisons with the proposed DASEIN method, we consider the following three benchmarks:

\begin{itemize}
	\item[1)]
	\textbf{Test-d}: Direct deployment of the trained source domain DNN models to the target domain, i.e., setting $ \theta^{\mathcal{T}} = \theta^{\mathcal{S}} $.
	\item[2)]
	\textbf{SC-DA} \cite{a28}: The semantic coding network is first trained with the source domain dataset. Then, it trains GAN-based DNNs to transform the target dataset into the form of the source domain dataset. After transformation, the target domain data is fed into the trained source domain semantic encoding network for subsequent classification. 
	\item[3)]
	\textbf{DANN} \cite{a22}: Domain adaptation neural network consists of a feature extractor, a label predictor, and a domain classifier. The adversarial interaction between the domain classifier and the feature extractor enables the feature extractor to learn domain-invariant features.
\end{itemize}\par

\begin{figure}[t]
	\centering
	\subfloat[train]{\includegraphics[width=0.7\hsize, height=0.66\hsize]{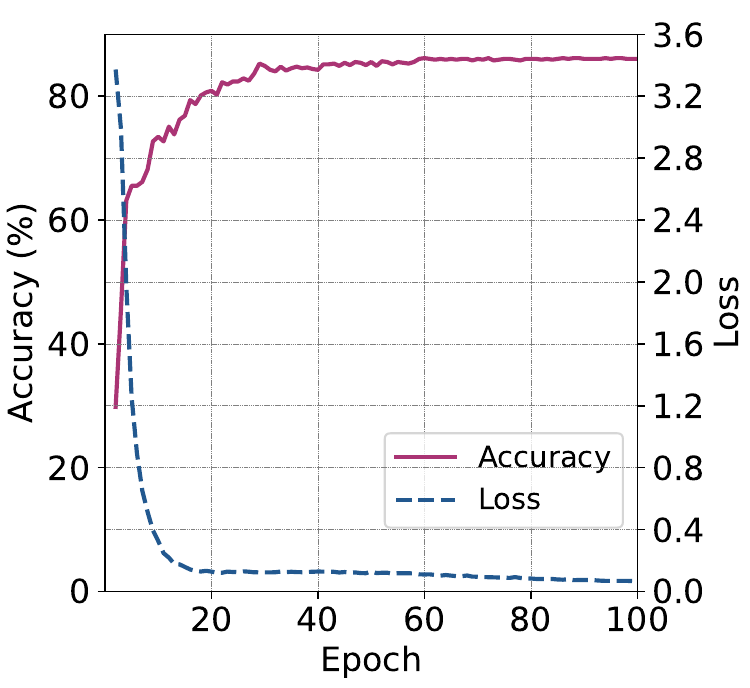}\label{train}}
	\hfill
	\subfloat[finetune]{\includegraphics[width=0.7\hsize,height=0.66\hsize]{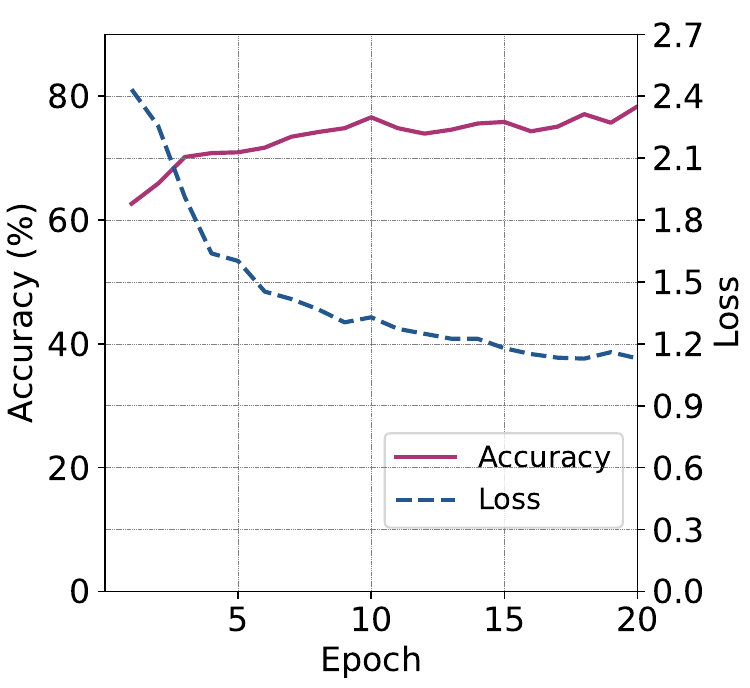}\label{finetune}}
	\caption{The accuracy curve and loss curve of DASEIN for transfer task A→W: (a) SNR = 5 dB and CR = 0.1; (b) SNR = -17 dB and CR = 0.1.}
	\label{fig:trend}
\end{figure}

\subsection{Analog Experiment Result\label{IV-B}}
\subsubsection{Tne Effectiveness of UDA\label{IV-B1}}

\begin{table}[t]
	\centering
	\caption{OFFICE-31: Accuracy ($\%$) Comparisons of Different Methods Applied to Different Transfer Tasks When $ CR = 0.1 $ and $ SNR = 5 dB $.}
	\scriptsize 
	\renewcommand{\arraystretch}{1.5} % 设置行高为1.5倍
	\begin{tabular}{ccccc}
		\toprule
		Task & Test-d & SC-DA & DANN & DASEIN-S1 \\
		\hline
		A→W & $ 65.98\pm0.84 $ & $ 31.66\pm0.87 $ & $ 77.91\pm0.67 $ & \bm{$ 85.11\pm0.07 $} \\
		D→W & $ 88.06\pm0.59 $ & $ 75.03\pm0.85 $ & $ 95.38\pm0.34 $ & \bm{$ 97.65\pm0.16 $} \\
		W→D & $ 94.80\pm0.93 $ & $ 79.76\pm1.33 $ & $ 95.69\pm0.59 $ & \bm{$ 99.78\pm0.06 $} \\
		A→D & $ 68.19\pm0.93 $ & $ 33.65\pm0.98 $ & $ 71.88\pm1.16 $ & \bm{$ 84.44\pm0.47 $} \\
		D→A & $ 52.73\pm0.39 $ & $ 22.28\pm0.52 $ & $ 59.45\pm0.38 $ & \bm{$ 63.40\pm0.19 $} \\
		W→A & $ 51.90\pm0.38 $ & $ 20.39\pm0.55 $ & $ 59.10\pm0.40 $ & \bm{$ 62.15\pm0.13 $} \\
		\hline
		Avg & $ 70.28 $ & $ 43.79 $ & $ 76.57 $ & \bm{$ 82.09 $} \\\hline
	\end{tabular}
	\label{tab:acc1}
\end{table}

\begin{table}[t]
	\centering
	\caption{VLCS: Accuracy ($\%$) Comparisons of Different Methods Applied to Different Transfer Tasks When $ CR = 0.1 $ and $ SNR = 5 dB $.}
	\scriptsize 
	\renewcommand{\arraystretch}{1.5} % 设置行高为1.5倍
	\begin{tabular}{ccccc}
		\toprule
		Task & Test-d & SC-DA & DANN & DASEIN-S1 \\
		\hline
		L→S & $ 48.98\pm0.58 $ & $ 43.68\pm0.32 $ & $ 51.40\pm0.25 $ & \bm{$ 64.56\pm0.15 $} \\
		L→V & $ 58.67\pm0.47 $ & $ 45.90\pm0.55 $ & $ 59.68\pm0.17 $ & \bm{$ 65.06\pm0.24 $} \\
		S→L & $ 59.58\pm0.50 $ & $ 45.05\pm0.80 $ & $ 63.11\pm0.23 $ & \bm{$ 67.03\pm0.11 $} \\
		S→V & $ 58.95\pm0.47 $ & $ 49.52\pm0.31 $ & $ 61.56\pm0.30 $ & \bm{$ 61.74\pm0.25 $} \\
		V→L & $ 61.13\pm0.38 $ & $ 48.40\pm0.34 $ & $ 64.44\pm0.29 $ & \bm{$ 65.86\pm0.28 $} \\
		V→S & $ 70.40\pm0.67 $ & $ 50.85\pm0.75 $ & $ 72.40\pm0.32 $ & \bm{$ 76.46\pm0.22 $} \\
		\hline
		Avg & $ 59.62 $ & $ 47.23 $ & $ 62.10 $ & \bm{$ 66.79 $} \\\hline
	\end{tabular}
	\label{tab:acc1_vlcs}
\end{table}

\begin{figure*}[t]
	\centering
	\subfloat[Test-d]{\includegraphics[width=0.40\hsize, height=0.36\hsize]{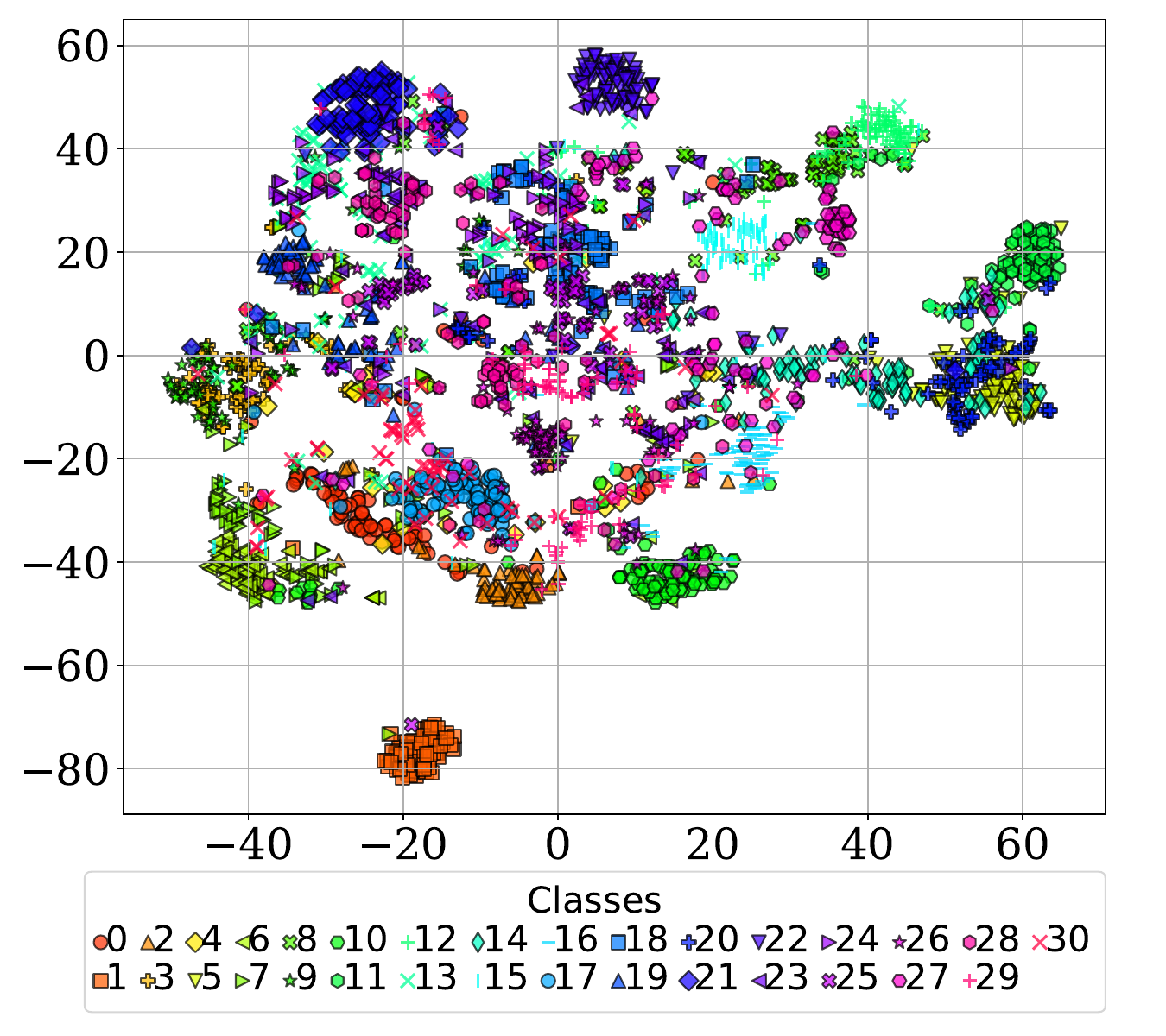}\label{tsne_test-d}}
	\hspace{0.05\linewidth}  % 调整这里的间距
	\subfloat[SC-DA]{\includegraphics[width=0.40\hsize, height=0.36\hsize]{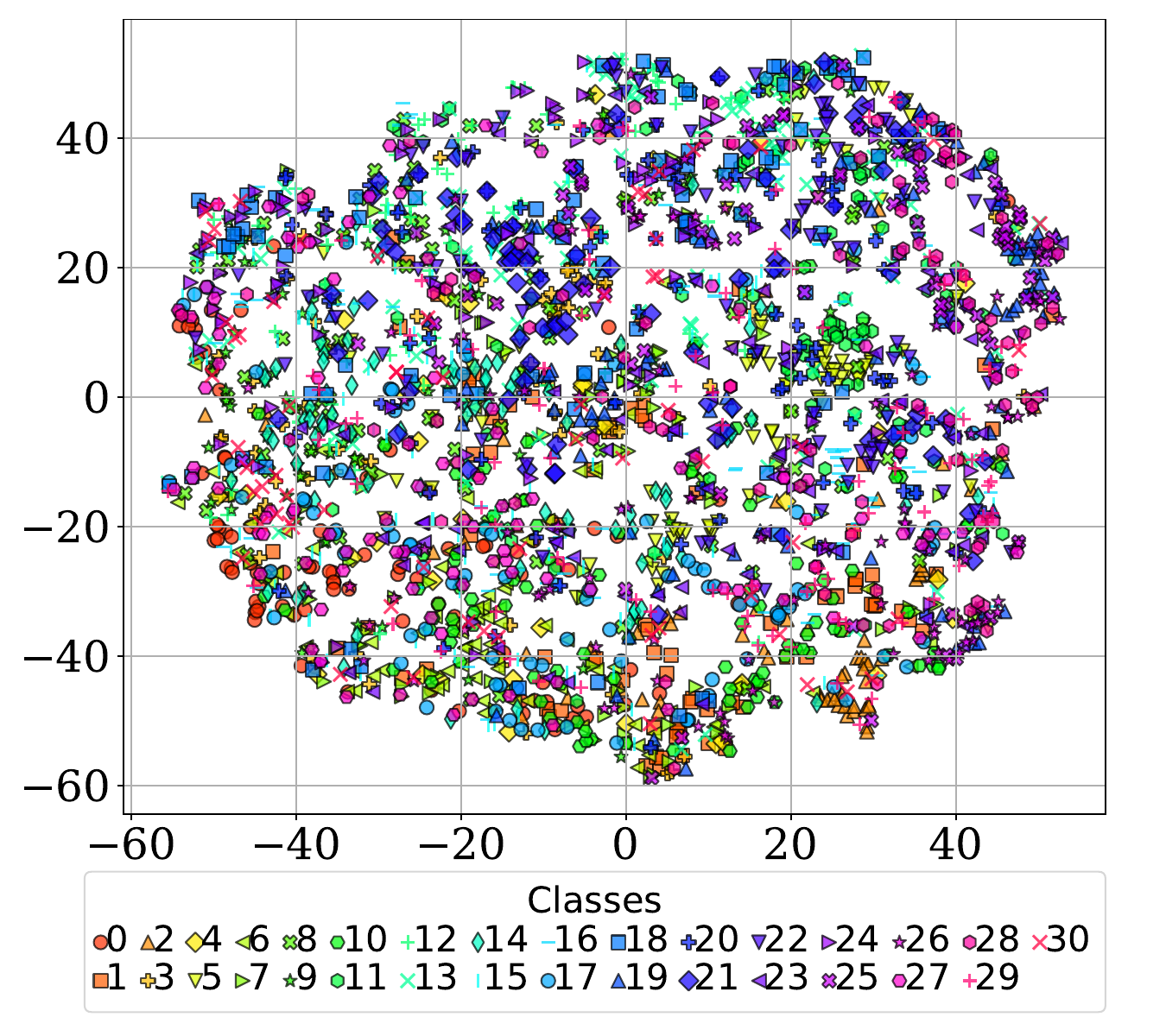}\label{tsne_SC-DA}}
	\hspace{0.05\linewidth}  % 调整这里的间距
	\subfloat[DANN]{\includegraphics[width=0.40\hsize, height=0.36\hsize]{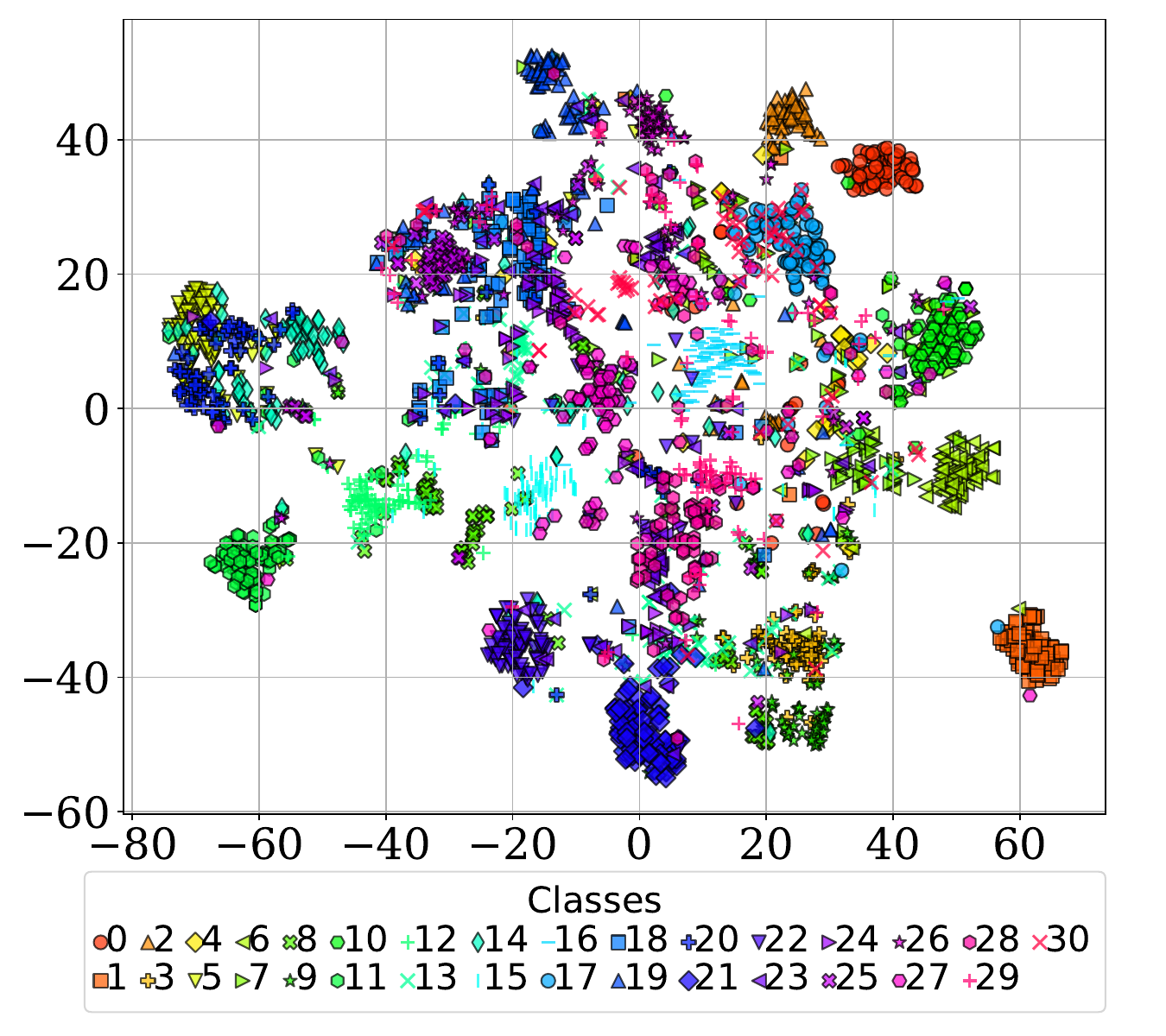}\label{tsne_DANN}}
	\hspace{0.05\linewidth}  % 调整这里的间距
	\subfloat[DASEIN-S1]{\includegraphics[width=0.40\hsize, height=0.36\hsize]{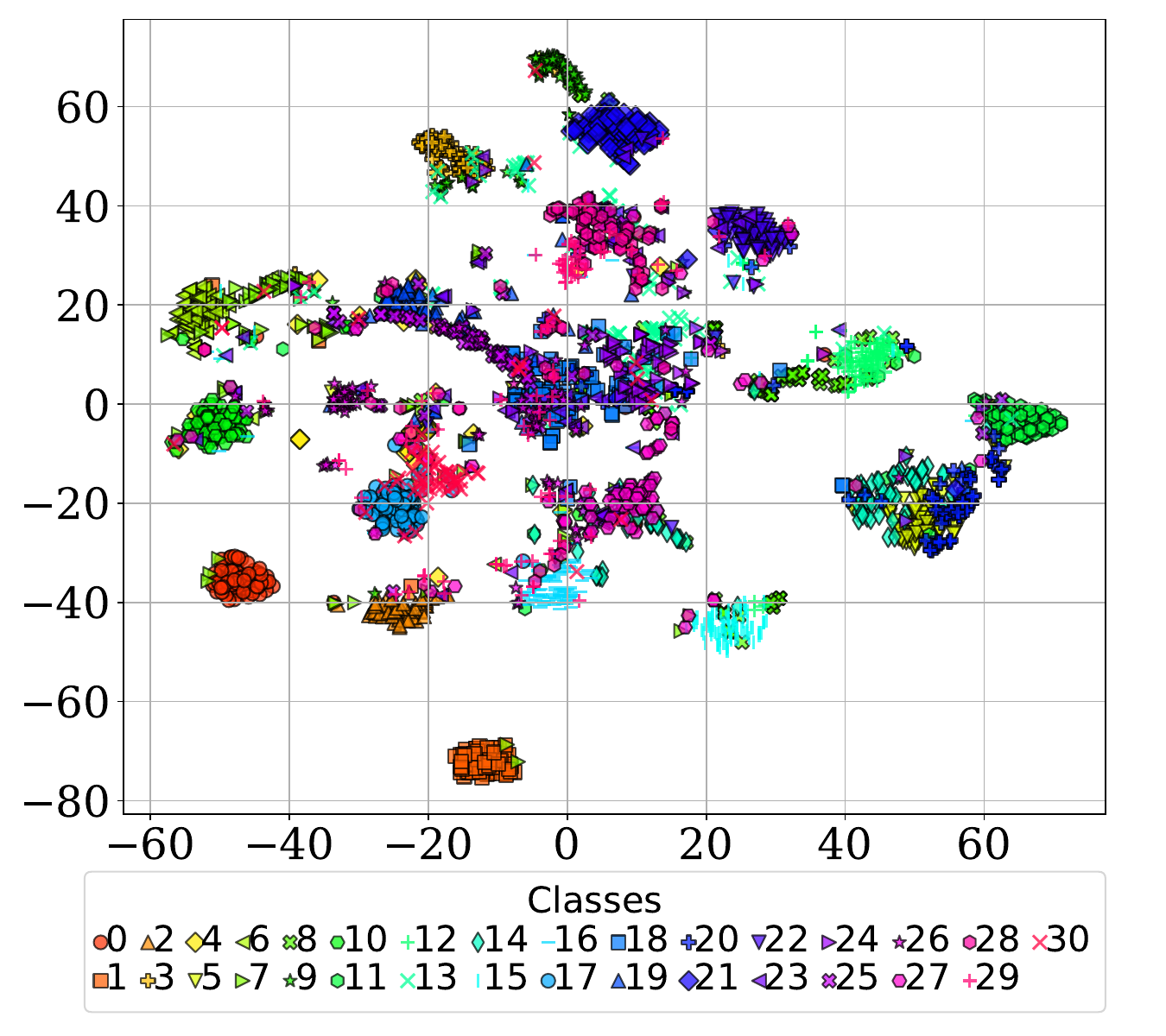}\label{tsne_DASEIN}}
	\caption{Visualization of the target domain feature distribution obtained using different methods in the W→A task.}
	\label{tsne}
\end{figure*}

\begin{figure*}[t]
	\centering
	\subfloat[Test-d]{\includegraphics[width=0.25\hsize, height=0.22\hsize]{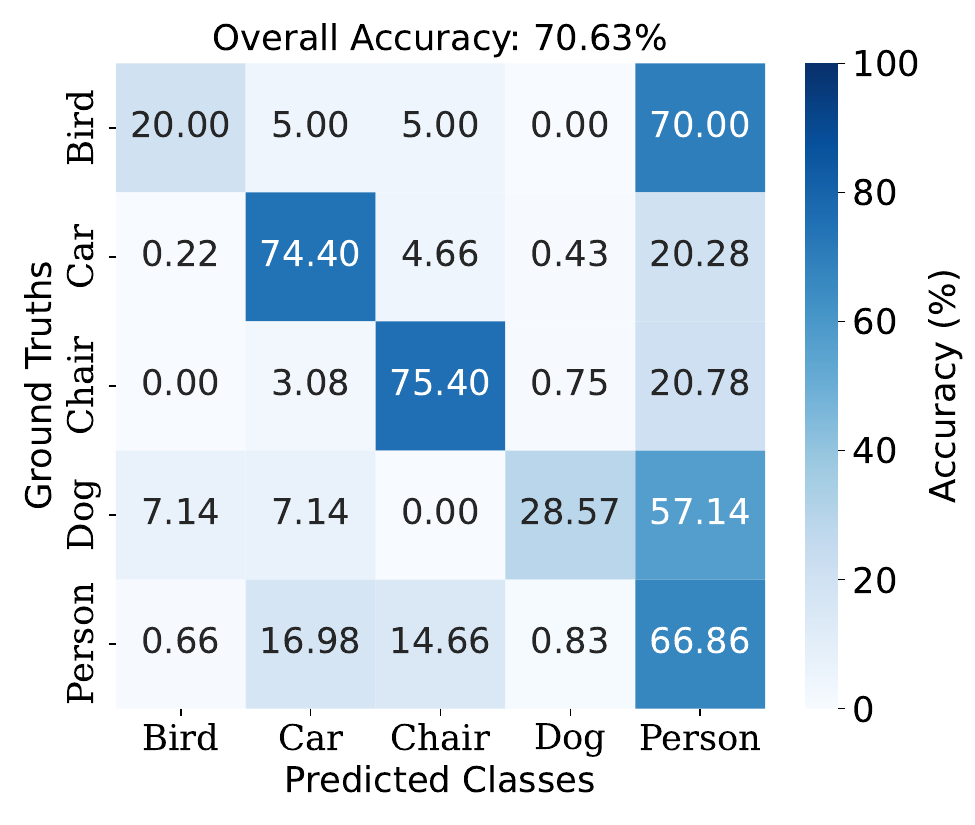}\label{matrix_test-d}}
	\subfloat[SC-DA]{\includegraphics[width=0.25\hsize, height=0.22\hsize]{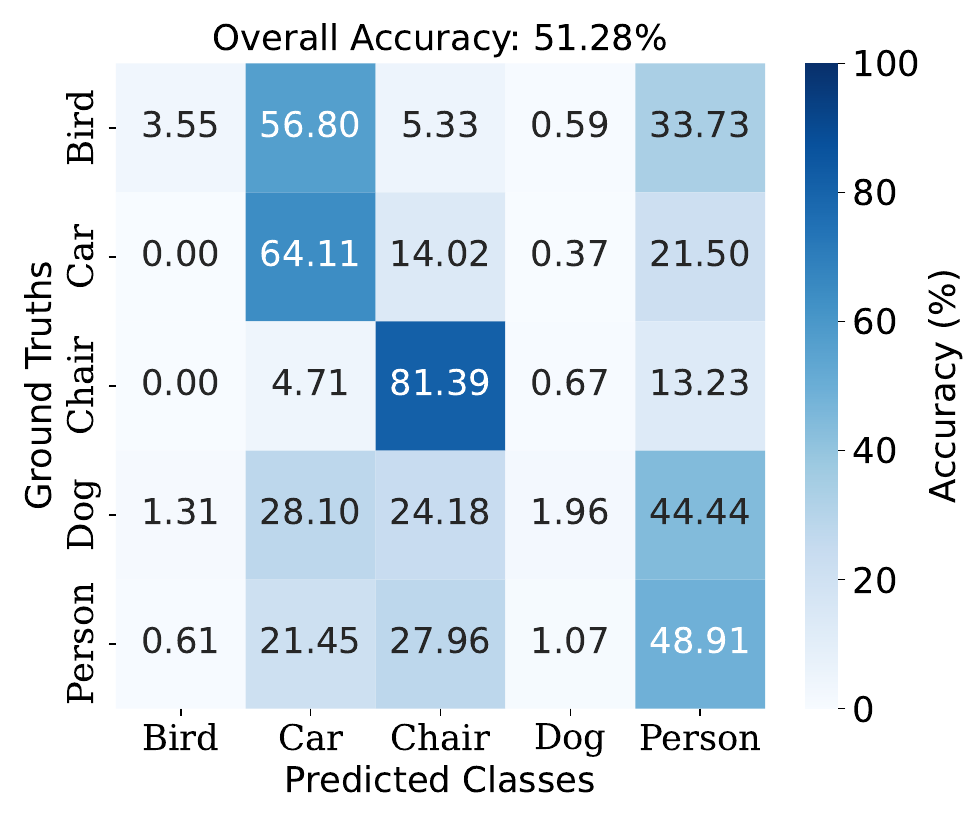}\label{matrix_SC-DA}}
	\subfloat[DANN]{\includegraphics[width=0.25\hsize, height=0.22\hsize]{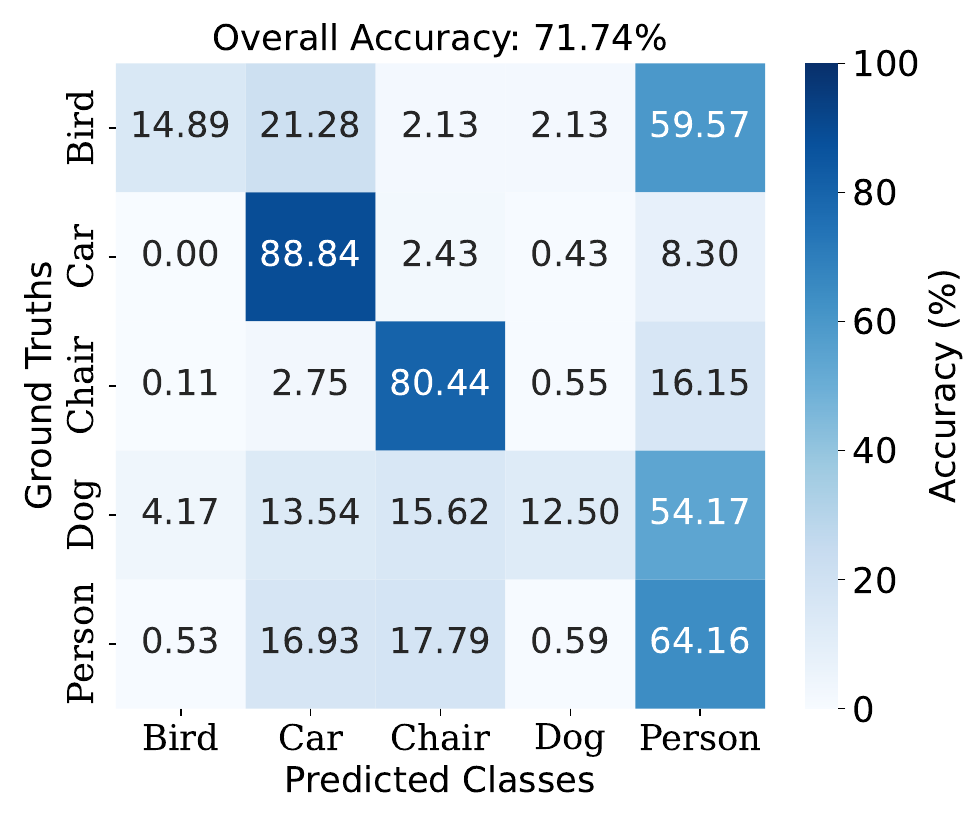}\label{matrix_DANN}}
	\subfloat[DASEIN-S1]{\includegraphics[width=0.25\hsize, height=0.22\hsize]{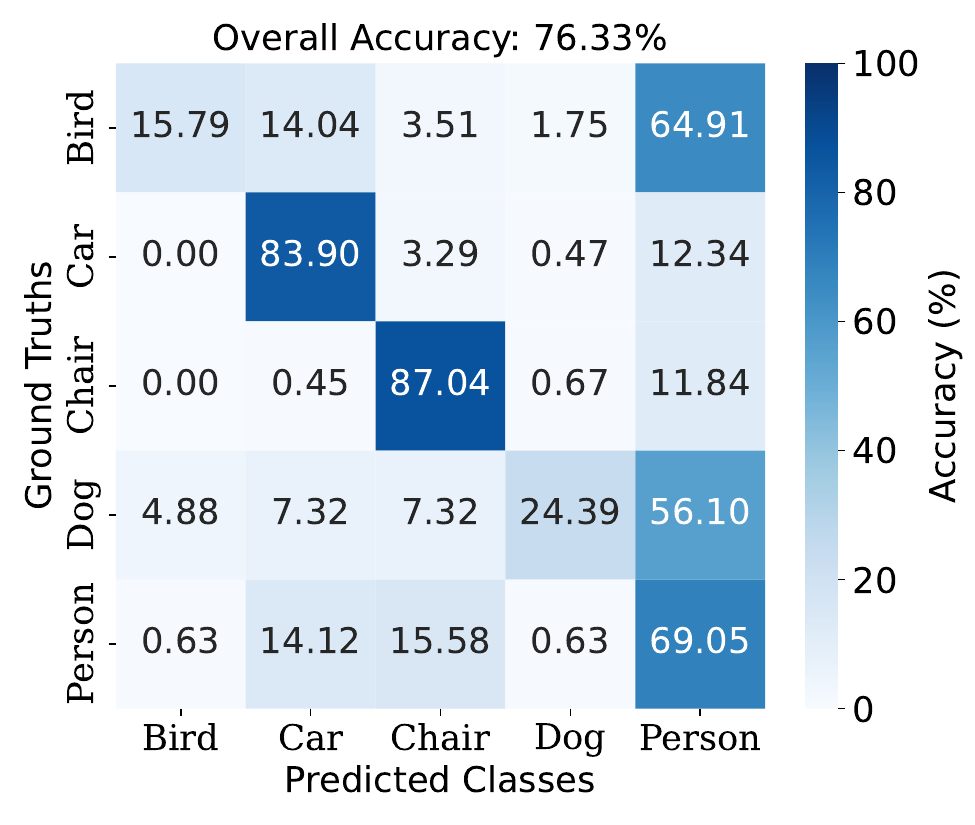}\label{matrix_DASEIN}}
	\caption{Confusion matrices of the target domain obtained using different methods in the V→S task.}
	\label{matrix}
\end{figure*}

To show the effectiveness of the devised UDA method in handling data distribution variation, we assume the target and source domains have the same channel condition. In particular, we set CR = 0.1, and the SNR for all the links is 5 dB. In this case, we only need step 1 of DASEIN, and we refer to the method as DASEIN-S1. Fig. \ref{train} shows the accuracy and training loss of DASEIN-S1 for the A→W transfer task. We can see that the test accuracy quickly converges in around 30 epochs. \par
Table \ref{tab:acc1} shows the inference accuracy of 6 different cross-domain tasks on the OFFICE-31 dataset when different methods are used. Recall that W and D are of similar data distribution but of different resolutions, the transfer deployment performances of all the schemes are satisfactory for W→D and D→W tasks, e.g., DASEIN-S1 achieves 99.78$ \% $ accuracy of the W→D task. Therefore, we mainly focus on the performance of the more challenging transfer tasks between dataset A and \{W, D\}. For the tasks A→W and A→D, the proposed DASEIN-S1 achieves 85.11$ \% $ and 84.44$ \% $ accuracy. For the D→A and W→A tasks, DASEIN-S1 achieves 63.40$\%$ and 62.15$\%$ accuracy because the source domain D or W have much fewer images than A. In all four cases, the proposed DASEIN-S1 achieves the best performance, where it outperforms the three benchmark methods on average by 11.81$\%$, 38.30$\%$, and 5.52$\%$, respectively. Fig. \ref{tsne} shows the t-SNE \cite{a33} projection of the feature distribution of the target domain data after applying different UDA methods in the W→A task, mapped into a two-dimensional space. Each point represents a sample, with color and shape indicating different classes. We observe that, compared to other methods, DASEIN is more effective in clustering samples of the same class together, such as class 9 (green star-shaped points), while achieving clear separation between different classes. This demonstrates the effectiveness of DASEIN in achieving high classification accuracy when source distribution changes significantly. \par
Table \ref{tab:acc1_vlcs} shows the inference accuracy of 6 cross-domain tasks on the VLCS dataset when different methods are used. DASEIN-S1 outperforms the three benchmark methods in all transfer tasks. On average, our method outperforms the three benchmark methods by 7.17$\%$, 19.56$\%$, and 4.69$\%$, respectively. As shown in Fig. \ref{matrix}, we present the confusion matrices of the target domain obtained by various methods in the V→S task. The results indicate that all methods achieve higher accuracy in recognizing categories such as ``car" and ``chair" but perform poorly in identifying ``bird" and ``dog." This phenomenon can be attributed to the label heterogeneity in the VLCS dataset, where the sample sizes of ``bird" and ``dog" are relatively small, causing the model to be biased toward predicting categories with larger sample sizes, such as ``car," ``chair," and ``person." Despite these challenges, our method still outperforms the comparative approaches. This verifies the proposed UDA procedure in handling data distribution variation of edge inference system deployment. Since the channel condition variations are unrelated to the data, we consider only the OFFICE-31 dataset in the following experiments to avoid repetitions. \par

It is worth mentioning that SC-DA has the worst performance and even performs worse than Test-d without domain adaptation. In fact, it is reported in \cite{a28} that SC-DA has excellent performance for transferring tasks with ample data samples, e.g., from MNIST to SVHN datasets with 60000 training samples. This is mainly because SC-DA uses GAN to directly generate target domain images from source domain images. However, generating complex images with GANs typically requires a large amount of training data to effectively capture image features. Due to the limited size of the available dataset, the GAN-based SC-DA method fails to generate effective images in our experiment. Therefore, we do not consider it in the following analysis. \par

\subsubsection{The Effectiveness of DASEIN\label{IV-B2}}

\begin{figure}[t]
	\centering
	\subfloat[CR=0.1]{\includegraphics[width=0.68\hsize, height=0.66\hsize]{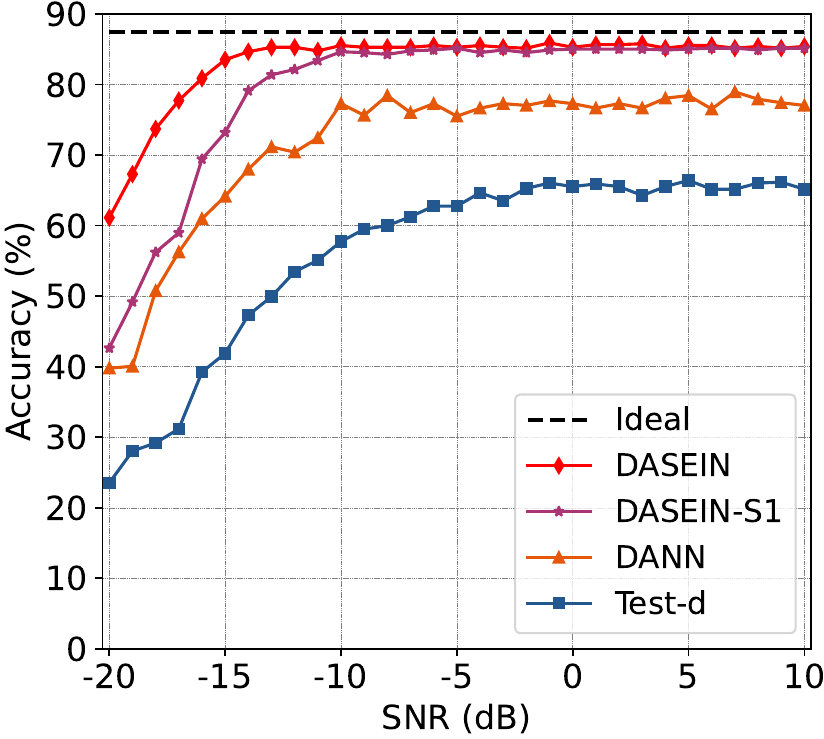}\label{cr204}}
	\hfill
	\subfloat[CR=0.25]{\includegraphics[width=0.68\hsize,height=0.66\hsize]{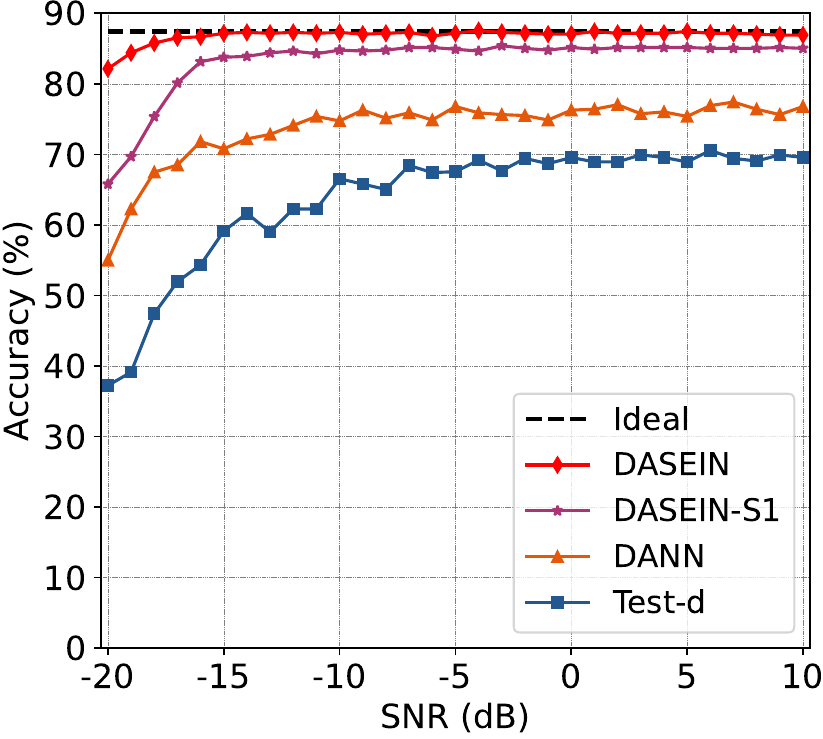}\label{cr512}}
	\caption{The impact of CR and SNR on the accuracy of a transfer task A→W: (a) CR = 0.1; (b) CR = 0.25.}
	\label{fig:ex2_Figure_1}
\end{figure}

We continue to evaluate the proposed DASEIN in tackling variations of both data and channel distribution. For data distribution variation, we consider the A→W task because we can collect sufficient data samples in the source domain dataset A. Besides, we consider SNR = 5 dB for all the links in the source domain. In Fig. \ref{fig:ex2_Figure_1}, we plot the inference accuracy of different methods under varying SNR and two different CRs. Here, the ideal benchmark refers to applying the proposed DASEIN with no SNR variation in the target domain and CR = 1. We can see that the inference accuracies of all the methods deteriorate at a lower SNR. Still, the proposed DASEIN has the best performance among all the methods considered, especially when the SNR of the target domain is below -10 dB. Fig. \ref{finetune} shows the accuracy curve and loss curve of the proposed DASEIN after finetuning the model when SNR = -17 dB and CR = 0.1. We observe an evident improvement in the task accuracy after fewer epochs of finetuning using the KD method, e.g., SNR = -20 dB, it outperforms benchmarks DASEIN-S1, DANN, and Test-d by 18.49$\%$, 21.33$\%$, and 37.61$\%$, respectively. At high SNR, e.g., larger than -5 dB, the performance improvement as a result of finetuning becomes marginal. When SNR = 5 dB in the target domain, DASEIN outperforms benchmarks DASEIN-S1, DANN, and Test-d by 0.5$\%$, 7.09$\%$, and 19.11$\%$, respectively.\par

We can also observe the impact of CR on the performance of the transfer tasks. Consider an extremely low SNR = -20 dB at the target domain, the performance gap between DASEIN and the ideal case is 23.29$ \% $ when CR = 0.1; however, the gap shrinks to only 5.28$ \% $ when we increase CR to 0.25. This indicates that a larger CR, and thus a larger feature dimension, attains higher capability against more noisy measurements. This also gives us a design insight that we should set different training methods based on the target domain channel condition $ \bm{\sigma} $. For instance, when the target domain has a good channel condition, e.g., SNR is greater than -10 dB, we only need to apply the UDA to align the data distributions (i.e., the DASEIN-S1 method). Meanwhile, under a bad channel condition, e.g., SNR $ \in $ [-15, 10] dB, we can effectively improve the accuracy through model fine-tuning with KD (i.e., the complete DASEIN method). However, as the SNR drops to a very low value, e.g., -20 dB, finetuning should be combined with an increased CR to achieve a consistently high inference accuracy. \par

\subsubsection{Bandwidth-accuracy Tradeoff\label{IV-B3}}
\begin{figure}[t]%调节图片位置，h：浮动；t：顶部；b:底部；p：当前位置
	\centering
	\includegraphics[width=0.86\linewidth]{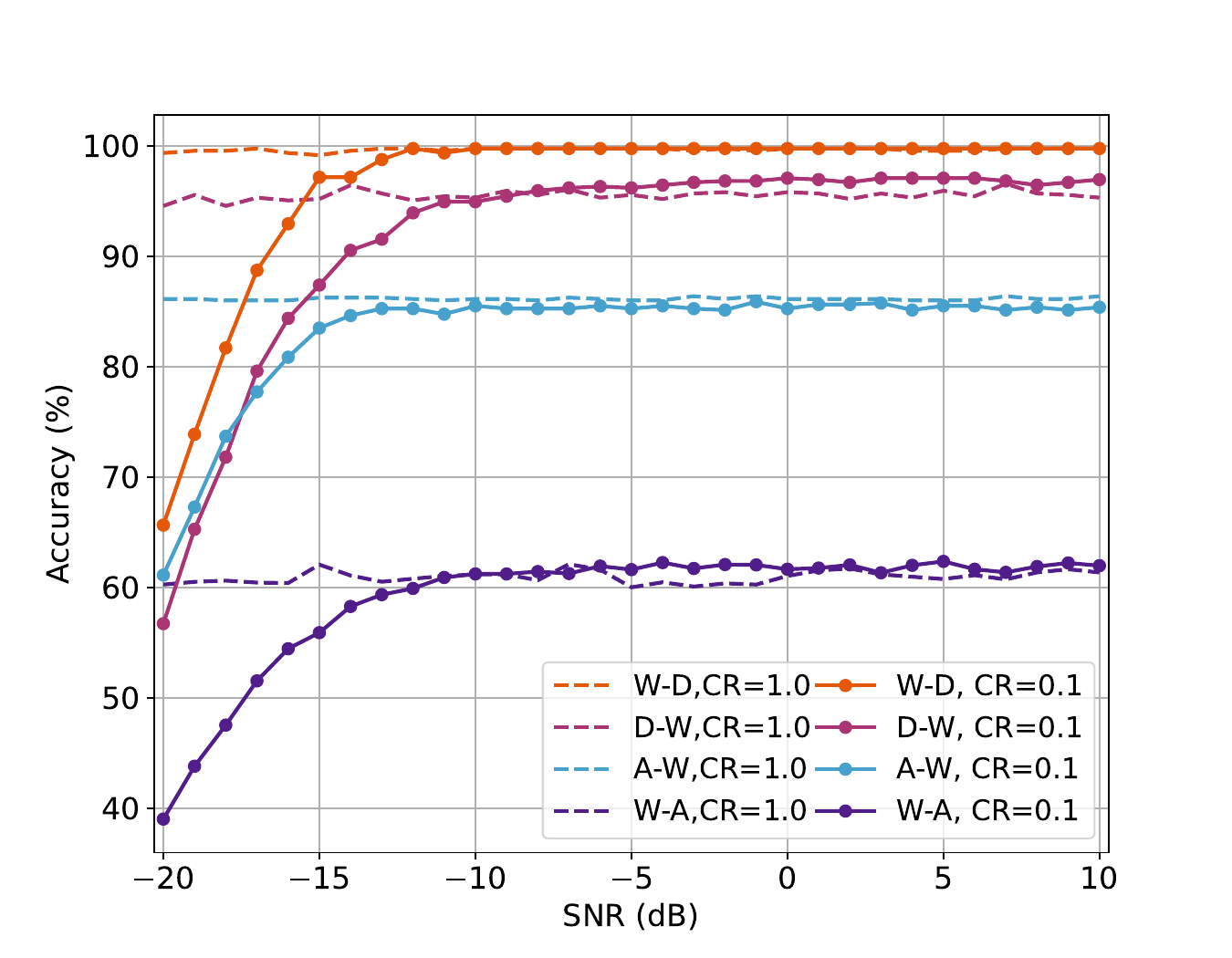}% 中括号中的为调节图片大小
	\caption{Impact of CR on the performance of DASEIN under both data and channel variations.}
	\label{fig:ex4}%文中引用该图片代号
\end{figure}

\begin{figure}[t]%调节图片位置，h：浮动；t：顶部；b:底部；p：当前位置
	\centering
	\includegraphics[width=0.9\linewidth]{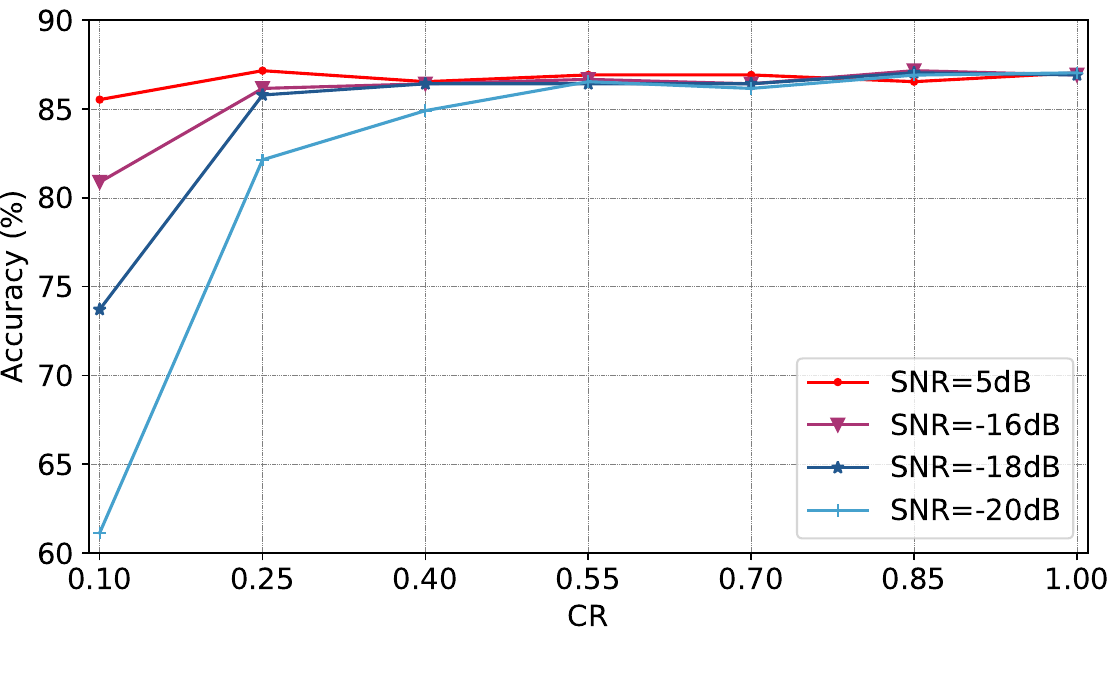}% 中括号中的为调节图片大小 1.0, 0.15
	\caption{Impact of CR on the task A→W performance of DASEIN under both data and channel variations.}
	\label{fig:ex4_cr_acc}%文中引用该图片代号
\end{figure}

In Fig. \ref{fig:ex4}, we study the impact of the CR on the proposed DASEIN in handling different joint variations of data (different transfer tasks) and channel (different SNRs) distributions. The style difference between the W and D images is minimal, with the main distinction being at the pixel level. For simplicity of exposition, we use the experimental results of the four transfer tasks W→D, D→W, A→W, and W→A to analyze in Fig. \ref{fig:ex4}. The dashed and solid lines indicate the results when CR = 1 and 0.1, respectively. When CR = 1, for all 4 transfer tasks, we observe that the performance of DASEIN is consistent in all SNRs considered, indicating a strong noise-resistant capability under sufficient redundancy in the feature space. With a high CR, however, the system also consumes excessive communication resources to transmit the features of large dimensions. As we emphasize more on communication efficiency by reducing CR to 0.1, we observe comparable performance to the case with CR = 1 when SNR is higher than -10 dB, but severely degraded inference accuracy as SNR drops below -10 dB. The above results show that we need to consider a bandwidth-accuracy performance tradeoff on setting the value of CR under a low SNR in the target domain. Next, we examine the A→W task to further analyze the tradeoffs between bandwidth and accuracy for DASEIN in Fig. \ref{fig:ex4_cr_acc}. We find that increasing CR at low SNR (e.g., less than -16 dB) can significantly enhance the inference accuracy (e.g., by more than $20\%$). However, the improvement becomes marginal under good channel condition (e.g., above 5 dB). \par
The above experiment results demonstrate the efficiency of DASEIN in handling the variations of data and channel distributions in extensive transfer deployment scenarios of edge inference systems. Besides, we have also concluded from the results with some interesting design insight, where we should flexibly perform model fine-tuning and adjust the feature dimension based on the SNR of the target domain deployment scenario. When the channel condition in the target domain is bad, e.g., SNR $ < $ -10 dB, increasing the CR to 0.55 can effectively enhance inference performance. Conversely, when the target domain has a good channel condition, e.g., SNR $ > $ -10 dB, the CR can be appropriately reduced to minimize the amount of data that needs to be transmitted.

\subsection{Performance of Digital Communication\label{IV2}}
\subsubsection{Setup\label{IV2-A}}
For the digital transmission experiments, we use the OFFICE-31 dataset. We set $ r $ = 3 in the differentiation function \eqref{e_9}. Besides, we reset $ E_f $ = 10 in Algorithm \ref{alg:a2}. The rest of the experiment setup is the same as the above analog scheme. Unless otherwise stated, set qb = 2 bit, CR =0.1, and the SNR for all the links is 5 dB.\par 

\begin{table}[t]
	\centering
	\caption{OFFICE-31: Accuracy ($\%$) Comparisons of Different Methods Applied to Different Transfer Tasks in Digital Communication When $ q_b $ = 2$ bit $, CR = 0.1, and SNR = 5$ dB $ .}
	\scriptsize 
	\renewcommand{\arraystretch}{1.5} % 设置行高为1.5倍
	\begin{tabular}{ccccc}
		\toprule
		Task & Test-d & SC-DA & DANN & DASEIN-S1 \\
		\hline
		A→W & $ 68.33\pm0.58 $ & $ 33.67\pm1.32 $ & $ 74.92\pm0.85 $ & \bm{$ 84.54\pm0.17 $}  \\
		D→W & $ 92.72\pm0.73 $ & $ 81.31\pm1.20 $ & $ 97.54\pm0.60 $ & \bm{$ 98.32\pm0.06 $}  \\
		W→D & $ 97.98\pm0.44 $ & $ 85.02\pm1.20 $ & $ 99.32\pm0.32 $ & \bm{$ 100.00\pm0.00 $} \\
		A→D & $ 73.25\pm0.62 $ & $ 37.90\pm1.25 $ & $ 77.90\pm0.76 $ & \bm{$ 84.74\pm0.13 $}  \\
		D→A & $ 57.56\pm0.30 $ & $ 19.11\pm0.54 $ & $ 62.46\pm0.19 $ & \bm{$ 66.32\pm0.24 $}  \\
		W→A & $ 55.02\pm0.32 $ & $ 12.28\pm0.81 $ & $ 61.54\pm0.51 $ & \bm{$ 65.40\pm0.14 $}  \\
		\hline
		Avg & $ 74.11 $ & $ 44.88 $ & $ 78.95 $ & \bm{$ 83.22 $} \\\hline
	\end{tabular}
	\label{tab:acc2}
\end{table}

\begin{figure}[t]
	\centering
	\subfloat[CR=0.1]{\includegraphics[width=0.68\hsize, height=0.66\hsize]{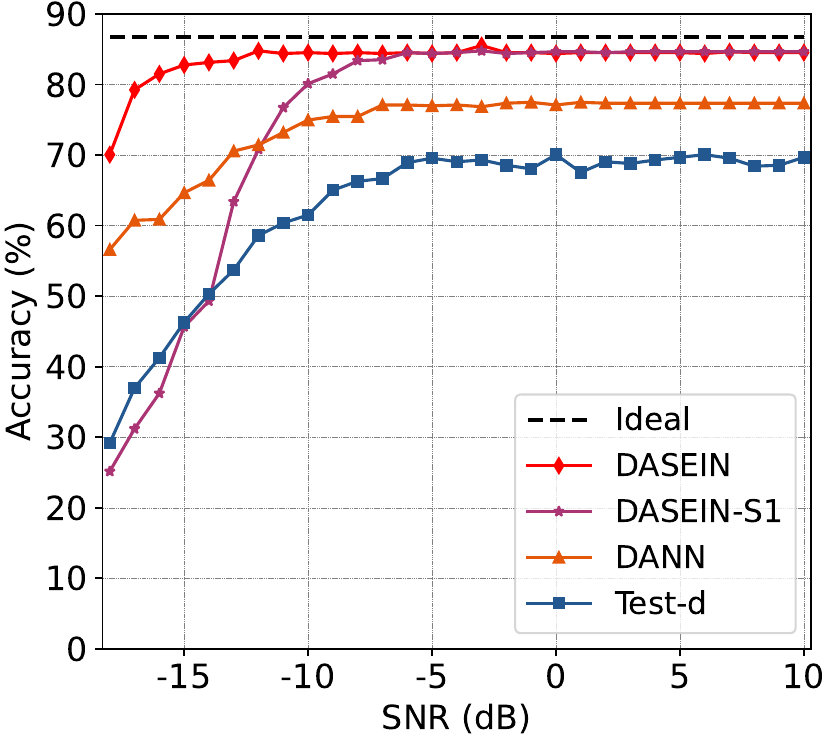}\label{cr204_d}}
	\hfill
	\subfloat[CR=0.25]{\includegraphics[width=0.68\hsize,height=0.66\hsize]{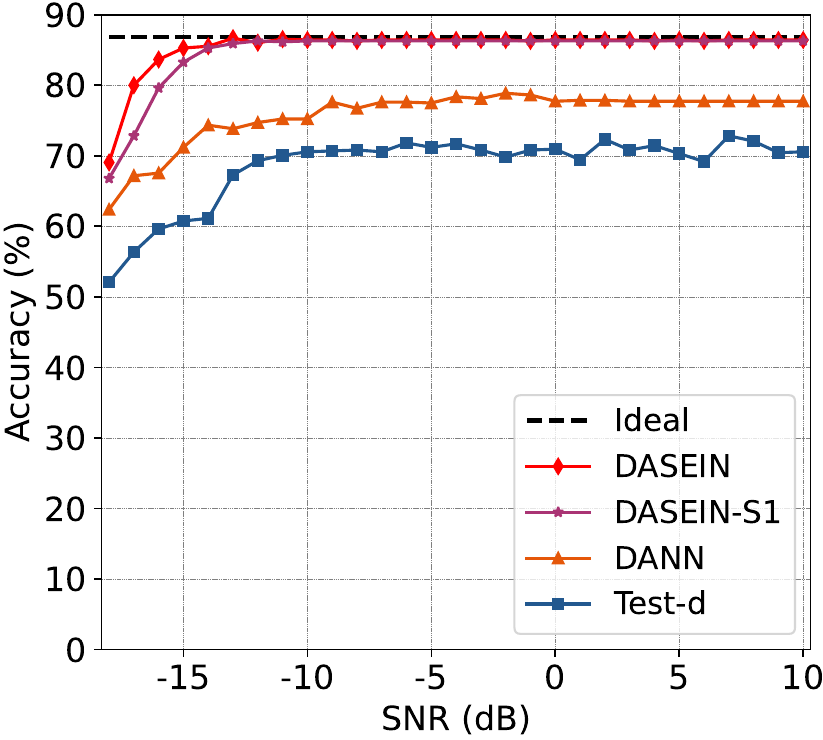}\label{cr512_d}}
	\caption{The impact of CR and SNR on the accuracy of a transfer task A→W when $ q_b $ = 2bit: (a) CR = 0.1; (b) CR = 0.25.}
	\label{fig:ex2_Figure_1_d}
\end{figure}

\subsubsection{Result\label{IV2-B}}
Table \ref{tab:acc2} presents the inference accuracy of 6 different cross-domain tasks under different methods. DASEIN-S1 achieves 100$ \% $ and 98.32$ \% $ accuracy in the W→D and D→W tasks, respectively. For the tasks A→W and A→D, DASEIN-S1 achieves 84.54$ \% $ and 84.74$ \% $ accuracy. For the D→A and W→A tasks, DASEIN-S1 achieves 66.32$\%$ and 65.40$\%$ accuracy. In all four cases, the proposed DASEIN-S1 achieves the best performance, where it outperforms the three benchmark methods on average by 9.11$\%$, 38.34$\%$, and 4.27$\%$, respectively. The results of digital scheme are similar as those in Table \ref{tab:acc1}, while the digital scheme slightly outperforms the analog scheme in most experiments, thanks to the noise-resilient capability of digital modulations. \par

We continue to evaluate the effectiveness of the proposed DASEIN in dealing with joint data-channel distribution variations. Fig. \ref{fig:ex2_Figure_1_d} shows that the inference accuracy of all the methods decreases at lower SNR. However, the proposed DASEIN performs the best among all the methods, especially when the SNR in the target domain is lower than -10 dB. We observe that DASEIN outperforms the benchmarks DASEIN-S1, DANN, and Test-d by 44.90$\%$, 13.46$\%$, and 40.88$\%$ respectively at low SNRs, e.g., SNR = -18 dB. When SNR = 5 dB in the target domain, DASEIN outperforms benchmarks DASEIN-S1, DANN, and Test-d by 0.12$\%$, 7.29$\%$, and 14.96$\%$, respectively. Meanwhile, by increasing the CR from $0.1$ to $0.25$, we observe evident increase of inference accuracy in the low SNR conditions. \par

\begin{figure}[t]%调节图片位置，h：浮动；t：顶部；b:底部；p：当前位置
	\centering
	\includegraphics[width=0.9\linewidth,height=0.46\hsize]{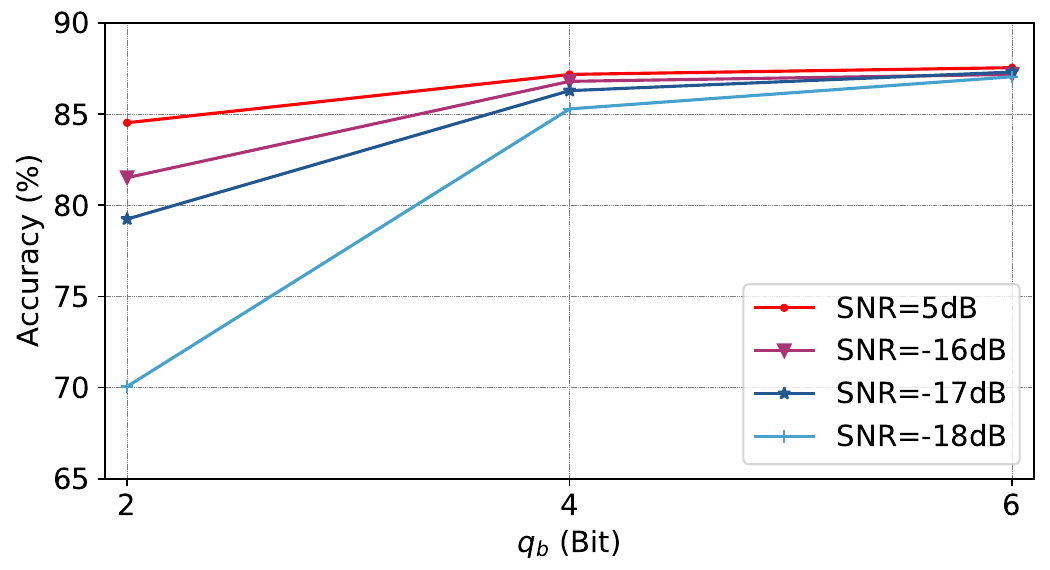}% 中括号中的为调节图片大小
	\caption{The impact of $ q_b $ on the accuracy of a transfer task A→W under both data and channel variations.}
	\label{fig:bit_acc}%文中引用该图片代号
\end{figure}

In Fig. \ref{fig:bit_acc}, we investigate the impact of $ q_b $ on the performance of the proposed DASEIN in handling variations of data and channel distributions for task A→W. A larger $ q_b $ indicates higher resolution in transmitting the analog feature, however, at the cost of higher bandwidth consumption. Evidently, a larger $ q_b $ leads to higher inference accuracy in all the SNR conditions considered. However, the improvement is significantly only under low SNR, e.g., more than $14\%$ when SNR=-18 dB, while marginal under favorable channel condition, e.g., less than $2\%$ when SNR = 5 dB. \par 

The experimental results demonstrate that DASEIN can derive effective design strategies for digital communication. When the channel conditions in the target domain are favorable, e.g., SNR is greater than -10 dB, applying DASEIN-S1 alone is sufficient to resist distribution shifts. However, under more challenging conditions, e.g., SNR $ \in $ [-15, 10] dB, using the complete DASEIN approach can significantly improve accuracy. Furthermore, when the SNR drops to an extremely low level, e.g., -18 dB, $ q_b $ should be reinforced to ensure stable and high inference accuracy.\par

\section{Conclusions}\label{V}

In that paper, we introduced a label-free transferable deployment method DASEIN for edge inference systems with different data and channel distributions. It consists of an UDA step to handle the data distribution variation and a subsequent KD fine-tuning step to address the difference in channel distributions. Experiment results have verified the effectiveness of DASEIN in maintaining high inference accuracy in different unfamiliar deployment scenarios, and revealed a bandwidth-accuracy performance tradeoff in the transfer deployment of edge inference system. \par
We conclude the paper with some interesting future working directions. First, we consider a Gaussian channel in this paper to illustrate the knowledge transfer process under different edge inference setups. In fact, the proposed method is agnostic to the channel distribution, and can be applied to wireless fading channel conditions, such as Rayleigh and Rician channels. Given the average SNR of a fixed edge inference system, we can simulate the fading channel condition following the specific distribution in the training process. In this case, the proposed DASEIN model can adapt to wireless fading channels of different average SNRs under dissimilar deployment setups. Second, this paper considers a static sensor deployment. In practice, the sensing device can be mobile, such as cameras mounted on drones. In this case, the channel SNRs between the sensing device and the edge server vary rapidly. To achieve a balanced performance between communication efficiency and inference accuracy, it requires to design a dynamic transmission rate adaptation scheme to adjust the dimension of the transit features.

\appendices
\section{The Gaussian kernel function} \label{A}

The expression of the Gaussian kernel function is: 
\begin{equation}\label{e13}
	\mathcal{K}(x_1, x_2) =  e^{-\frac{\| x_1 - x_1 \|^2}{2 \sigma_b ^2}}.
\end{equation}
We can rewrite \eqref{e13} and expand it according to Taylor series as:
\begin{equation}\label{e14}
	\begin{aligned}
		&\mathcal{K}(x_1, x_2) = e^{-\frac{( x_1 - x_2 )^2}{2 \sigma_b ^2}} = e^{-\frac{x_1^2 + x_2^2}{2 \sigma_b ^2}}e^{\frac{x_1 x_2}{\sigma_b ^2}}\\
		&= e^{-\frac{x_1^2 + x_2^2}{2 \sigma_b ^2}} \left( 1 + \frac{1}{1!}\frac{x_1 x_2}{\sigma_b ^2} + \cdots + \frac{1}{n!} \left( \frac{x_1 x_2}{\sigma_b ^2} \right) ^n + \cdots \right) \\
		&= e^{-\frac{x_1^2 + x_2^2}{2 \sigma_b ^2}} \left( 1\cdot 1 + \frac{1}{1!}\frac{x_1}{\sigma_b}\frac{x_2}{\sigma_b} + \cdots + \frac{1}{n!}\frac{x_1^n}{\sigma_b^n}\frac{x_2^n}{\sigma_b^n} + \cdots \right) \\
		&= \phi(x_1) \phi(x_2).
	\end{aligned}
\end{equation}
From \eqref{e14}, we can derive the expression for the mapping function $ \phi(x) $:
\begin{equation}\label{e15}
	\phi(x) = e^{-\frac{x^2}{2 \sigma_b ^2}} \left( 1, \sqrt{\frac{1}{1!}}\frac{x}{\sigma_b}, \cdots, \sqrt{\frac{1}{n!}}\frac{x^n}{{\sigma_b}^n}, \cdots \right).
\end{equation}

\bibliographystyle{IEEEtran}
\bibliography{ref2.bib}

\begin{thebibliography}{10}
\providecommand{\url}[1]{#1}
\csname url@samestyle\endcsname
\providecommand{\newblock}{\relax}
\providecommand{\bibinfo}[2]{#2}
\providecommand{\BIBentrySTDinterwordspacing}{\spaceskip=0pt\relax}
\providecommand{\BIBentryALTinterwordstretchfactor}{4}
\providecommand{\BIBentryALTinterwordspacing}{\spaceskip=\fontdimen2\font plus
\BIBentryALTinterwordstretchfactor\fontdimen3\font minus
  \fontdimen4\font\relax}
\providecommand{\BIBforeignlanguage}[2]{{%
\expandafter\ifx\csname l@#1\endcsname\relax
\typeout{** WARNING: IEEEtran.bst: No hyphenation pattern has been}%
\typeout{** loaded for the language `#1'. Using the pattern for}%
\typeout{** the default language instead.}%
\else
\language=\csname l@#1\endcsname
\fi
#2}}
\providecommand{\BIBdecl}{\relax}
\BIBdecl

\bibitem{a1}
C.~Zeng, J.-B. Wang, M.~Xiao, C.~Ding, Y.~Chen, H.~Yu, and J.~Wang,
  ``Task-oriented semantic communication over rate splitting enabled wireless
  control systems for urllc services,'' \emph{IEEE Trans. Commun.}, vol.~72,
  no.~2, pp. 722--739, Feb. 2024.

\bibitem{a2}
J.~Huang, D.~Li, C.~Huang, X.~Qin, and W.~Zhang, ``Joint task and data oriented
  semantic communications: A deep separate source-channel coding scheme,''
  \emph{IEEE Internet Things J.}, vol.~11, no.~2, pp. 2255--2272, Jan. 2024.

\bibitem{a3}
X.~Peng, Z.~Qin, X.~Tao, J.~Lu, and L.~Hanzo, ``A robust semantic text
  communication system,'' \emph{IEEE Trans. Wireless Commun.}, vol.~23, no.~9,
  pp. 11\,372--11\,385, Apr. 2024.

\bibitem{a6}
J.~Yan, S.~Bi, and Y.-J.~A. Zhang, ``Optimal model placement and online model
  splitting for device-edge co-inference,'' \emph{IEEE Trans. Wireless
  Commun.}, vol.~21, no.~10, pp. 8354--8367, Oct. 2022.

\bibitem{a7}
H.~Zhang, H.~Wang, Y.~Li, K.~Long, and A.~Nallanathan, ``Drl-driven dynamic
  resource allocation for task-oriented semantic communication,'' \emph{IEEE
  Trans. Commun.}, vol.~71, no.~7, pp. 3992--4004, Jul. 2023.

\bibitem{a7_1}
J.~Du, J.~Wang, A.~Sun, J.~Qu, J.~Zhang, C.~Wu, and D.~Niyato, ``Joint
  optimization in blockchain- and mec-enabled space–air–ground integrated
  networks,'' \emph{IEEE Internet Things J.}, vol.~11, no.~19, pp.
  31\,862--31\,877, Jnl. 2024.

\bibitem{a7_2}
X.~Li and S.~Bi, ``Optimal {AI} model splitting and resource allocation for
  device-edge co-inference in multi-user wireless sensing systems,'' \emph{IEEE
  Trans. Wireless Commun.}, vol.~23, no.~9, pp. 11\,094--11\,108, Mar. 2024.

\bibitem{a7_3}
X.~Li, S.~Bi, H.~Zeng, B.~Lin, and L.~Xiaohui., ``Collaborative task offloading
  and resource allocation optimization for intelligent edge devices,''
  \emph{Chinese Journal on Internet of Things}, vol.~6, no.~4, pp. 41--52, Dec.
  2022.

\bibitem{a13_1}
N.-N. Dao, D.-N. Vu, Y.~Lee, S.~Cho, C.~Cho, and H.~Kim, ``Pattern-identified
  online task scheduling in multitier edge computing for industrial iot
  services,'' \emph{Mobile Information Systems.}, vol. 2018, no.~1, p. 2101206,
  Apr. 2018.

\bibitem{a13_2}
C.~Cai, X.~Yuan, and Y.-J.~A. Zhang, ``Multi-device task-oriented communication
  via maximal coding rate reduction,'' \emph{IEEE Trans. Wireless Commun.},
  Sept. 2024.

\bibitem{a13_3}
X.~Li, S.~Bi, S.~Wang, X.~Li, and Y.-J.~A. Zhang, ``Digital semantic
  device-edge co-inference with task-oriented arq,'' \emph{IEEE Trans. Veh.
  Technol.}, vol.~73, no.~9, pp. 13\,986--13\,990, Apr. 2024.

\bibitem{a8}
J.~Shao, Y.~Mao, and J.~Zhang, ``Task-oriented communication for multidevice
  cooperative edge inference,'' \emph{IEEE Trans. Wireless Commun.}, vol.~22,
  no.~1, pp. 73--87, Jul. 2023.

\bibitem{a9}
N.~Patwa, N.~Ahuja, S.~Somayazulu, O.~Tickoo, S.~Varadarajan, and S.~Koolagudi,
  ``Semantic-preserving image compression,'' in \emph{Proc. IEEE Int. Conf.
  Image Process (ICIP).}, United Arab Emirates, Oct. 2020, pp. 1281--1285.

\bibitem{a9_1}
X.~Li, S.~Bi, S.~Wang, X.~Li, and Y.-J.~A. Zhang, ``Digital semantic
  device-edge co-inference with task-oriented arq,'' \emph{IEEE Trans. Veh.
  Technol.}, vol.~73, no.~9, pp. 13\,986--13\,990, Sept. 2024.

\bibitem{a9_2}
P.~Li, G.~Cheng, J.~Kang, R.~Yu, L.~Qian, Y.~Wu, and D.~Niyato, ``Fast:
  Fidelity-adjustable semantic transmission over heterogeneous wireless
  networks,'' in \emph{Proc. IEEE Int. Conf. Commun. (ICC)}, Rome, Italy, May.
  2023, pp. 4689--4694.

\bibitem{a9_3}
C.~Cai, X.~Yuan, and Y.-J.~A. Zhang, ``End-to-end learning for task-oriented
  semantic communications over mimo channels: An information-theoretic
  framework,'' \emph{IEEE J. Sel. Areas Commun.}, vol.~43, no.~4, pp.
  1292--1307, Jan. 2025.

\bibitem{a10}
M.~U. Lokumarambage, V.~S.~S. Gowrisetty, H.~Rezaei, T.~Sivalingam,
  N.~Rajatheva, and A.~Fernando, ``Wireless end-to-end image transmission
  system using semantic communications,'' \emph{IEEE Access.}, vol.~11, pp.
  37\,149--37\,163, Apr. 2023.

\bibitem{a11}
J.~Shao, X.~Zhang, and J.~Zhang, ``Task-oriented communication for edge video
  analytics,'' \emph{IEEE Trans. Wireless Commun.}, vol.~23, no.~5, pp.
  4141--4154, May. 2024.

\bibitem{a12}
Z.~Weng and Z.~Qin, ``Semantic communication systems for speech transmission,''
  \emph{IEEE J. Sel. Areas Commun.}, vol.~39, no.~8, pp. 2434--2444, Aug. 2021.

\bibitem{a13}
L.~Yan, Z.~Qin, R.~Zhang, Y.~Li, and G.~Y. Li, ``Resource allocation for text
  semantic communications,'' \emph{IEEE Wireless Commun. Lett.}, vol.~11,
  no.~7, pp. 1394--1398, Jul. 2022.

\bibitem{a14}
E.~Beck, C.~Bockelmann, and A.~Dekorsy, ``Semantic information recovery in
  wireless networks,'' \emph{Sensors.}, vol.~23, no.~14, p. 6347, Mar. 2023.

\bibitem{a15}
M.~Ding, J.~Li, M.~Ma, and X.~Fan, ``Snr-adaptive deep joint source-channel
  coding for wireless image transmission,'' in \emph{Proc. IEEE Int. Conf.
  Acoust., Speech Signal Process (ICASSP).}, Toronto, Ontario, Canada, Jun.
  2021, pp. 1555--1559.

\bibitem{a16}
J.~Xu, B.~Ai, W.~Chen, A.~Yang, P.~Sun, and M.~Rodrigues, ``Wireless image
  transmission using deep source channel coding with attention modules,''
  \emph{IEEE Trans. Circuits Syst. Video Technol.}, vol.~32, no.~4, pp.
  2315--2328, Apr. 2022.

\bibitem{a17}
J.~Shao, Y.~Mao, and J.~Zhang, ``Learning task-oriented communication for edge
  inference: An information bottleneck approach,'' \emph{IEEE J. Sel. Areas
  Commun.}, vol.~40, no.~1, pp. 197--211, Jan. 2022.

\bibitem{a18}
S.~J. Pan and Q.~Yang, ``A survey on transfer learning,'' \emph{IEEE Trans.
  Knowl. Data Eng.}, vol.~22, no.~10, pp. 1345--1359, Oct. 2010.

\bibitem{a19}
Z.~Zhu, K.~Lin, A.~K. Jain, and J.~Zhou, ``Transfer learning in deep
  reinforcement learning: A survey,'' \emph{IEEE Trans. Pattern Anal. Mach.
  Intell.}, vol.~45, no.~11, pp. 13\,344--13\,362, Nov. 2023.

\bibitem{a20}
H.~Hou, S.~Bi, L.~Zheng, X.~Lin, Y.~Wu, and Z.~Quan, ``Dasecount:
  Domain-agnostic sample-efficient wireless indoor crowd counting via few-shot
  learning,'' \emph{IEEE Internet Things J.}, vol.~10, no.~8, pp. 7038--7050,
  Apr. 2023.

\bibitem{a21}
L.~Zheng, S.~Bi, S.~Wang, Z.~Quan, X.~Li, X.~Lin, and H.~Wang, ``Resmon:
  Domain-adaptive wireless respiration state monitoring via few-shot bayesian
  deep learning,'' \emph{IEEE Internet Things J.}, Dec. 2023.

\bibitem{a22}
Y.~Ganin, E.~Ustinova, H.~Ajakan, P.~Germain, H.~Larochelle, F.~Laviolette,
  M.~March, and V.~Lempitsky, ``Domain-adversarial training of neural
  networks,'' \emph{J. Mach. Learn. Res.}, vol.~17, no.~59, pp. 1--35, Apr.
  2016.

\bibitem{a23}
E.~Tzeng, J.~Hoffman, K.~Saenko, and T.~Darrell, ``Adversarial discriminative
  domain adaptation,'' in \emph{Proc. IEEE Int. Conf. Computer Vision and
  Pattern Recognition (CVPR).}, Hawaii, America, Jul. 2017, pp. 7167--7176.

\bibitem{a24}
P.~Oza, V.~A. Sindagi, V.~V. Sharmini, and V.~M. Patel, ``Unsupervised domain
  adaptation of object detectors: A survey,'' \emph{IEEE Trans. Pattern Anal.
  Mach. Intell.}, vol.~46, no.~6, pp. 4018--4040, Jun. 2024.

\bibitem{a25}
Y.~Zhu, F.~Zhuang, J.~Wang, G.~Ke, J.~Chen, J.~Bian, H.~Xiong, and Q.~He,
  ``Deep subdomain adaptation network for image classification,'' \emph{IEEE
  Trans. Neural Netw. Learn. Syst.}, vol.~32, no.~4, pp. 1713--1722, Apr. 2021.

\bibitem{a26}
H.~Xie, Z.~Qin, G.~Y. Li, and B.-H. Juang, ``Deep learning enabled semantic
  communication systems,'' \emph{IEEE Trans. Signal Process.}, vol.~69, pp.
  2663--2675, Apr. 2021.

\bibitem{a27}
Z.~Feng, D.~Cai, Z.~Liu, J.~Shan, and W.~Wang, ``Data adaptive semantic
  communication systems for intelligent tasks and image transmission,'' in
  \emph{Proc. Int. Conf. on AI-generated Content.}, Springer, Singapore, Nov.
  2023, pp. 105--117.

\bibitem{a281}
Y.~Sun, H.~Chen, X.~Xu, P.~Zhang, and S.~Cui, ``Semantic knowledge base-enabled
  zero-shot multi-level feature transmission optimization,'' \emph{IEEE Trans.
  Wireless Commun.}, vol.~23, no.~5, pp. 4904--4917, May. 2024.

\bibitem{a28}
H.~Zhang, S.~Shao, M.~Tao, X.~Bi, and K.~B. Letaief, ``Deep learning-enabled
  semantic communication systems with task-unaware transmitter and dynamic
  data,'' \emph{IEEE J. Sel. Areas Commun.}, vol.~41, no.~1, pp. 170--185, Jan.
  2023.

\bibitem{a29}
A.~Gretton, K.~M. Borgwardt, M.~J. Rasch, B.~Sch{\"o}lkopf, and A.~Smola, ``A
  kernel two-sample test,'' \emph{J. Mach. Learn. Res.}, vol.~13, no.~1, pp.
  723--773, Mar. 2012.

\bibitem{a29-1}
G.~Hinton, O.~Vinyals, and J.~Dean, ``Distilling the knowledge in a neural
  network,'' \emph{arXiv preprint arXiv:1503.02531}, 2015.

\bibitem{a29-2}
X.~Xu, M.~Li, C.~Tao, T.~Shen, R.~Cheng, J.~Li, C.~Xu, D.~Tao, and T.~Zhou, ``A
  survey on knowledge distillation of large language models,'' \emph{arXiv
  preprint arXiv:2402.13116}, 2024.

\bibitem{a30}
K.~He, X.~Zhang, S.~Ren, and J.~Sun, ``Deep residual learning for image
  recognition,'' in \emph{Proc. IEEE Int. Conf. Computer Vision and Pattern
  Recognition (CVPR).}, Las Vegas, NV, USA, Jun. 2016, pp. 770--778.

\bibitem{a30_1}
X.~Li, S.~Bi, S.~Wang, X.~Li, and Y.-J.~A. Zhang, ``Digital semantic
  device-edge co-inference with task-oriented arq,'' \emph{IEEE Trans. Veh.
  Technol.}, pp. 1--6, Apr. 2024.

\bibitem{a31}
K.~Saenko, B.~Kulis, M.~Fritz, and T.~Darrell, ``Adapting visual category
  models to new domains,'' in \emph{Proc. European Conference on Computer
  Vision (ECCV).}, Heraklion, Crete, Greece, Sept. 2010, pp. 213--226.

\bibitem{a32}
A.~Torralba and A.~A. Efros, ``Unbiased look at dataset bias,'' in \emph{Proc.
  IEEE Int. Conf. Computer Vision and Pattern Recognition (CVPR).}\hskip 1em
  plus 0.5em minus 0.4em\relax Colorado Springs, Colorado, USA: IEEE, June.
  2011, pp. 1521--1528.

\bibitem{a33}
L.~Van~der Maaten and G.~Hinton, ``Visualizing data using t-sne,'' \emph{J.
  Mach. Learn. Res.}, vol.~9, no.~11, Nov. 2008.

\end{thebibliography}

\end{document}